\pdfoutput=1
\documentclass[11pt]{article}
\usepackage[table,xcdraw]{xcolor}
\usepackage{acl}
\usepackage{subcaption}
\usepackage{xspace}
\usepackage{booktabs}
\usepackage{graphicx}
\usepackage{multirow}
\usepackage{tikz}
\usetikzlibrary{plotmarks}
\usepackage{fontawesome5}
\usepackage{tabularx}

\usepackage[normalem]{ulem}
\usepackage{times}
\usepackage{latexsym}
\usepackage{amsmath}
\usepackage{amsfonts}
\usepackage{tabularx}
\usepackage{cleveref}
\crefformat{section}{\S#2#1#3} % see manual of cleveref, section 8.2.1
\crefformat{subsection}{\S#2#1#3}
\crefformat{subsubsection}{\S#2#1#3}

\newcommand{\stitle}[1]{\vspace{1ex} \noindent{\bf #1.}}
\usepackage[T1]{fontenc}
\usepackage[utf8]{inputenc}

\usepackage{microtype}
\usepackage{array}
\usepackage{tabularx}
\usepackage{graphicx}

\usepackage{inconsolata}
\usepackage{algorithm}
\usepackage{algorithmic}
\usepackage{soul}
\usepackage{enumitem}
\usepackage{todonotes}
\definecolor{darkgreen}{rgb}{0.0, 0.5, 0.0}
\definecolor{carnelian}{rgb}{0.7, 0.11, 0.11}

  {\list{}{\leftmargin=#1\rightmargin=#1}\item[]}%
  {\endlist}

\usepackage{listings}
\title{Are Large Language Models Capable of Generating Human-Level Narratives?}
 \author{Yufei Tian$^{1}$\thanks{\ \ The two authors contributed equally.} \quad  Tenghao Huang$^{2*}$ \quad Miri Liu$^1$ \quad Derek Jiang$^1$ \quad Alexander Spangher$^2$ \\ \quad \textbf{Muhao Chen}$^{23}$  \quad \textbf{Jonathan May}$^2$ 
\quad \textbf{Nanyun Peng}$^1$ \\[7pt]
         $^1$University of California, Los Angeles, $^2$University of Southern California\\$^3$University of California, Davis
         \\[3pt]
         {\url{https://github.com/PlusLabNLP/Narrative-Discourse}} \\[5pt]
         {
         \texttt{yufeit@cs.ucla.edu} \quad \texttt{tenghaoh@usc.edu}
         }
         }

\begin{document}
\maketitle
\begin{abstract}

%As reliance on large language models (LLMs) for narrative creation grows, assessing their generated quality is crucial to understanding how they might impact our communications.
% 
This paper investigates the capability of LLMs in storytelling, focusing on narrative development and plot progression. 
We introduce a novel computational framework to analyze narratives through three discourse-level aspects: i) story arcs, ii) turning points, and iii) affective dimensions, including arousal and valence. By leveraging expert and automatic annotations, we uncover significant discrepancies between the LLM- and human- written stories. 
While human-written stories are suspenseful, arousing, and diverse in narrative structures, LLM stories are homogeneously positive and lack tension. Next, we measure narrative reasoning skills as a precursor to generative capacities, concluding that most LLMs fall short of human abilities in discourse understanding. Finally, we show that explicit integration of aforementioned discourse features can enhance storytelling, as is demonstrated by over 40\% improvement in neural storytelling in terms of diversity, suspense, and arousal. %Such advances promise to facilitate greater and more natural roles LLMs in human communication.

%%% Comments
% Audience have concerns that it is the default generation setting that makes it boring
% referring to figure on some terms
% start with we acknowledge LLMs can generate stories well. the question is how well they are. Previous eval/analysis focus on local features?
% objective description. highlight the difference --> don't argue it is good or bad
% describe in terms of metrics and objective numbers.
% can mention target audience - stories/movies
% human stories follow such arcs while LLM are flat
% mentioning "qualitative prior work" in this way make people feel we are following the prior work

%Leveraging insights from prior qualitative studies%that there exist gaps in story-telling ability in local-level discourse structures
\end{abstract}

\section{Introduction}
Storytelling serves as an integral part in shaping our understandings of ourselves, our society and our world \cite{langer1942philosophy, kaniss1991making}. As large language models (LLMs) grow in capabilities \cite{minaee2024large} and are integrated into our daily communicative routines \cite{kasneci2023chatgpt}, assessing the narrative structures of the stories they tell is crucial to understanding the ways they will shape our society.

Humans incorporate discourse structures that span local and global levels to captivate audiences, evoke emotions, convey complex messages, and share unique perspectives \cite{vonnegut1995shapes, van1980macrostructures}. %As described by \citet{van1980macrostructures}, humans incorporate discourse structures that span local and global %coherence levels to achieve the above purposes. 
A recent HCI study has pointed to gaps in machine storytelling ability at the global-level: despite being able to craft \textit{fluent} narratives, LLMs such as GPT-4 and Claude exhibit plot holes or produce repetitive themes that are less preferred by human critics \cite{chakrabarty2024art}. % development, such as , rather than word usage.\yufei{need to cite more related works?} 

\begin{figure}
    \centering
    \includegraphics[width=0.98\linewidth]{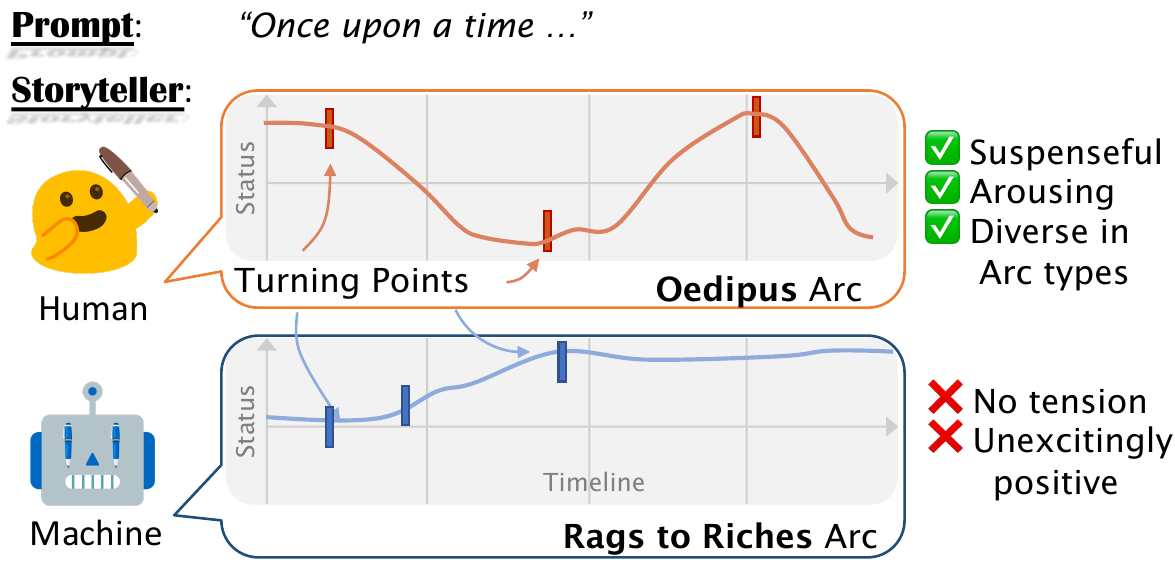}\vspace{-2mm}
    \caption{The story arc and turning point positions of human- and LLM- generated narratives. The vertical axis shows the character's fortune (bad to good), and the horizontal axis represents timeline (beginning to end). %LLMs tell less interesting stories than humans. 
    Compared with human storytellers, LLMs tend to (1) adopt homogeneously happier, less complex story arcs, (2) introduce plot turning points earlier in the timeline, and (3) have less suspense or fewer setbacks in their storylines. The impact of these differences grow as LLMs gain greater prominence in communicative patterns.}
    \label{fig:teaser}\vspace{-2mm}
\end{figure}

\begin{table*}[t]
\centering
\small
% \begin{tabularx}{\textwidth}{
    % m{0.7 cm}
    % *{7}{p{1.75cm}
% }}
\begin{tabular}{p{1.96cm}p{1.96cm}p{1.96cm}p{1.72cm}p{1.72cm}p{1.72cm}p{1.72cm}}
\toprule
  \textbf{Rags to Riches} &
  \textbf{Riches to Rags} &
  \textbf{Man in a Hole} &
  \textbf{Double Man in a Hole} &
  \textbf{Icarus} &
  \textbf{Cinderella} &
  \textbf{Oedipus} \\ 
  \midrule
% \textbf{
% \begin{tabular}[c]{@{}l@{}}Visual-\\ ization\end{tabular}
% }& 
   \includegraphics[width=1.72cm
   % 1.2\linewidth
   ]{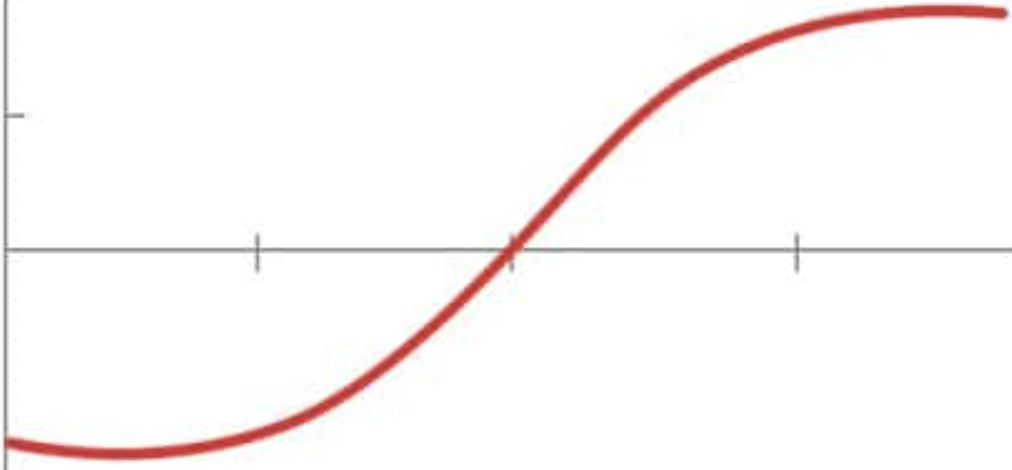}   &
   \includegraphics[width=1.72cm
   % 1.2\linewidth
   ]{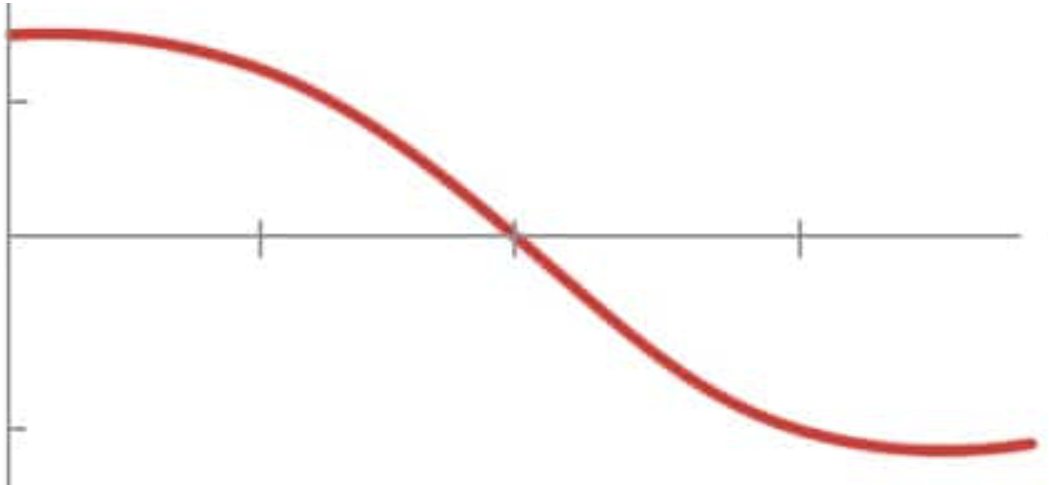} &
   \includegraphics[width=1.72cm
   % 1.2\linewidth
   ]{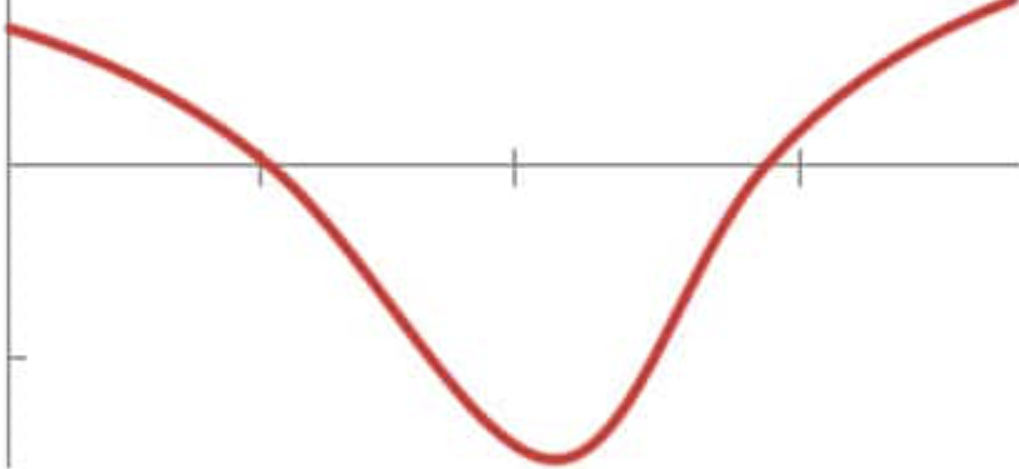} &
   \includegraphics[width=1.72cm
   % 1.2\linewidth
   ]{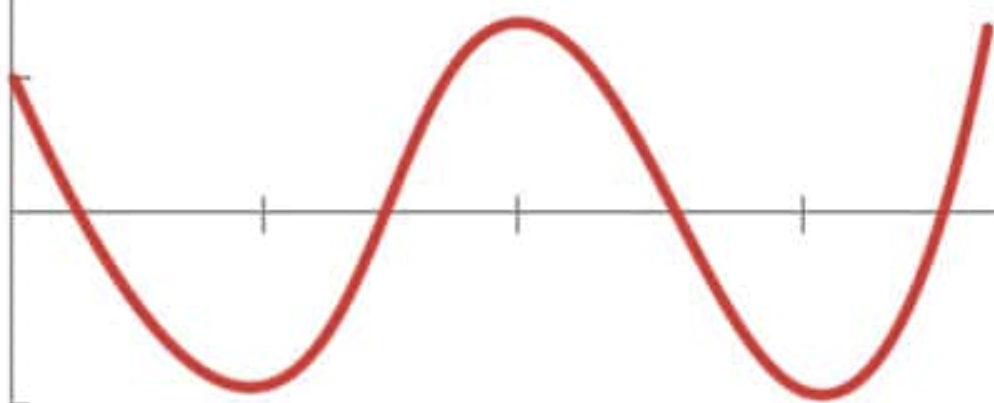} &
   \includegraphics[width=1.72cm
   % 1.2\linewidth
   ]{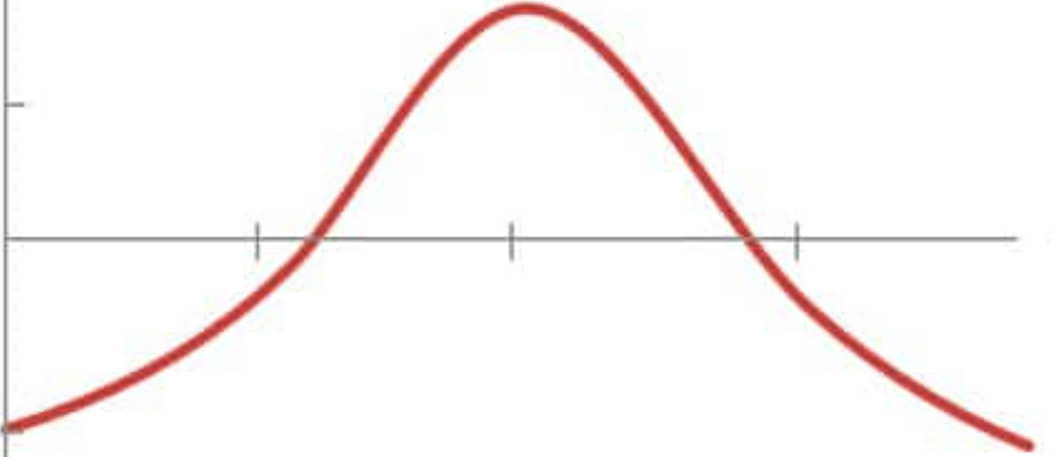} &
   \includegraphics[width=1.72cm
   % 1.2\linewidth
   ]{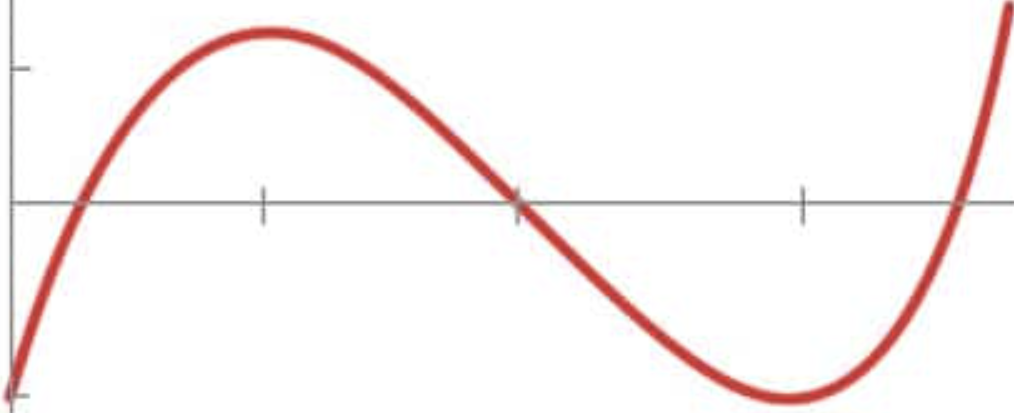}&
   \includegraphics[width=1.72cm
   % 1.2\linewidth
   ]{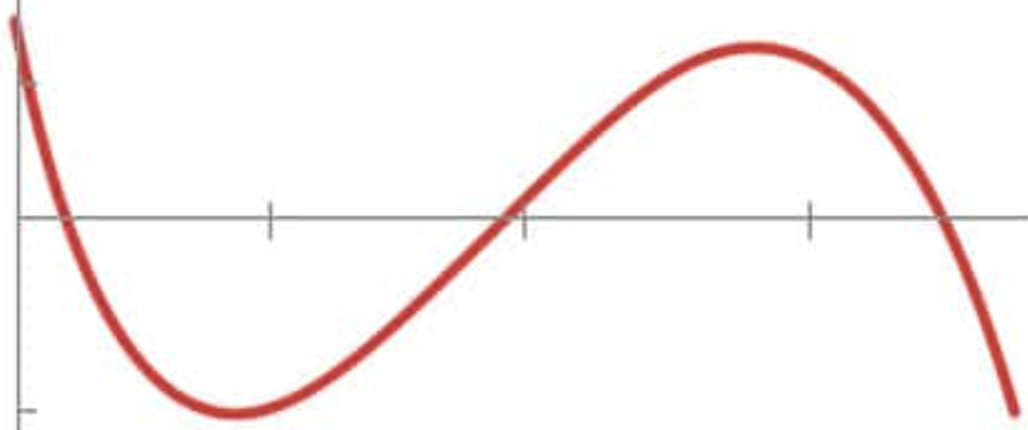}
   \\ 
   % \midrule
   \cmidrule(lr){1-1}\cmidrule(lr){2-2}\cmidrule(lr){3-3}\cmidrule(lr){4-4}\cmidrule(lr){5-5}\cmidrule(lr){6-6}\cmidrule(lr){7-7}
   
% \textbf{
% \begin{tabular}[c]{@{}l@{}}Descri-\\ ption\end{tabular}
% } &
  Starts low and gradually rises, ending in a high state. &
  Starts high and gradually falls, ending in a low state. &
  Starts high, has a dilemma or crisis and finally finds a way out. &
  Two cycles of fall and rise. &
  A rise followed by a sharp fall. &
  A rise, followed by a fall, ending with a significant rise. &
  A fall, followed by a rise, ending with a significant fall. \\ \bottomrule
% \end{tabularx}
\end{tabular}
\caption{\textbf{Story Arc Types:} We define and visualize the seven story arc types in our macro-level narrative discourse schema. Story arc types are derived from \citet{vonnegut1995shapes}, and are characterized by transformations of the story's protagonist(s) across the plot progression.}\label{tab:story_arc} \vspace{-2mm}
\end{table*}

%Left unanswered, however, are more fundamental questions about storytelling on a global coherence and comprehension level. 
However, a computational framework remains to be established for quantitative assessments of narratives at the global or discourse level.
%Drawing insights from qualitative story analysis \cite{vonnegut1995shapes}, 
We take a step towards the much-desired \textit{analytical framework} and attempt to answer pertinent questions such as: do stories generated by LLMs exhibit the same narrative complexity and diversity as human storytelling? Do LLMs have the capacity to comprehend narrative structures? Concretely, we measure narrative discourse structures at three distinct levels: 1) story arcs (\textit{i.e.,} macro-level narrative development), 2) turning points (\textit{i.e.,} meso-level shifts) and 3) arousal \& valence (\textit{i.e.,} micro-level dynamics). We collect a  dataset of movie synopses, on which we conduct a wide range of human and automated annotations for each of these levels, with the goals of (1) contrasting LLM and human storytelling, and (2) probing LLM narrative structure comprehension.

%  Do LLM-generated stories reach the same narrative complexity and diversity of human storytelling? Do LLMs comprehend narrative structures?

%\footnote{Our codes and the expert annotated dataset will be publicly released upon acceptance.}

First, we explore LLM storytelling abilities. As shown in Figure  \ref{fig:teaser} and  Section~\ref{sec:discourse_comparison}, LLMs such as GPT-4 exhibit notable deficiencies in \textit{narrative pacing}. These models often struggle to adequately develop critical turning points in a story, such as the major setback and climax, diminishing two key qualities for an engaging story: suspense and arousal. Additionally, machines are biased towards certain types of macro-level story-arcs and show a \textit{lack of narrative diversity}, particularly in avoiding negative plot progressions.

Next, we probe LLMs' \textit{narrative structure comprehension ability} (Section~\ref{sec:benchmark}) to test the hypothesis that poor comprehension affects LLMs' narrative generation abilities. % pointed to increased \textit{comprehension} affecting language modeling/generative abilities \cite{collobert2008unified, raffel2020exploring}; we benchmark narrative comprehension as a useful correlate to benchmarking narrative ability. 
To achieve this, we develop two benchmarks, (1) story arc identification and (2) turning point identification in stories, and evaluate Gemini Pro, Claude 3 Opus, Llama3, GPT-3.5 and GPT-4 \cite{reid2024gemini,anthropic2024claude,llama3modelcard,chatgpt2022,openai2023gpt4}. Again, we observe a substantial gap between most LLMs and human abilities, which matches our hypothesis. Interestingly, we find that the different discourse-levels reinforce each other: we can improve turning point identification including story arc information in the input, and vice versa.
 
Motivated by this finding, we explore whether we can improve machine story-telling %: \textit{Can we enhance LLM's storytelling abilities by leveraging the discourse features discussed earlier?} 
by leveraging the aforementioned discourse features, and we find promising results. Incorporating discourse reasoning in prompts can serve as important guidance towards better story generation (Section~\ref{sec:enhance}). In two parallel experiments, we demonstrate that integrating awareness of story arcs enhances model diversity (outperforming vanilla prompting by 45\%), whereas incorporating turning point information significantly improves narrative suspense and engagement (outperforming vanilla by 40\%). 

In summary, our contributions are threefold:

\begin{enumerate} [topsep=0pt, itemsep=-2pt, leftmargin=*]
    \item We unify three levels of discourse in narrative analysis: story arc, turning point, and affective dimension. Based on this, we present the first quantitative analysis framework to benchmark narrative development, and demonstrate %its practical applicability 
    that it can be operationalized by humans on benchmark dataset which we release (\S\ref{sec:background} and \S\ref{sec:data_collect_annotate}).
    \item We use this discourse framework to provide a novel comparison of LLM and human generative capacities by examining story-telling (\S\ref{sec:discourse_comparison}) and story-comprehension abilities (\S\ref{sec:benchmark}). We find that LLMs' abilities fall short of human abilities in both, but especially in story-telling. %abilities when evaluated on these discourse dimensions.
    \item We demonstrate that a discourse-aware generation process with LLMs (\S\ref{sec:enhance}) --- i.e. incorporating and reasoning about the story arc or turning points---enhances their overall narrative construction, as is reflected in improved suspense, emotion provocation, and narrative diversity. This lays the groundwork for future research to refine models to incorporate complex narrative structures for storytelling and beyond.
    %improvements can be made, LLMs still fall short of human performance, indicating a challenging new realm of evaluation.
\end{enumerate}

%Our work aims to advance the computational understanding of narratives through the lens of discourse elements. We hope the collected dataset and experimental results --- along with our proposed perspective --- will provide valuable insights into better automatic narrative generation and evaluation.  % \yufei{I wrote a forward-looking ending. Do we need to summarize our contributions again in bullet points?}

\section{Background: Discourse in Narratives}\label{sec:background}

%\yufei{This section talks about problem framing. 1) introduces What’s discourse-level features, and define TP, Story Arc. Briefly mention arousal and valence}\yufei{2) Why is discourse important for narratives? Motivate the the significance of TP and Story Arc from an NLP/ML perspective. Tp is importance features for Pacing, plot progression, and intensity. Arc is important for diversity and plot progression. Also say that TP4 and TP5 are most important/determinant TPs.}\yufei{Ideal length of this section: half page not including the figures}

We identify three aspects of plot progression in story-telling: %1. 
story-arcs (macro-level), 
turning-points (meso-level) and arousal/valence (micro-level), each representing a different level on which storytellers develop their narratives \cite{van1980macrostructures}. %We hypothesize that each level demonstrates a different story-telling capability.
We describe each of them before describing how we collect data to measure them in stories.

\begin{table*}[ht]
\centering
\small
\begin{tabular}{@{}>{\bfseries}m{4cm}m{12cm}@{}}
\toprule
Turning Point (TP) & \textbf{Description} \\ 
% \cmidrule(lr){1-1}\cmidrule(lr){2-2}
\midrule
TP1 - Opportunity & The introductory event that sets the stage for the narrative. \\
TP2 - Change of Plans & A pivotal moment where the main goal of the narrative is defined or altered. \\
TP3 - Point of No Return & The commitment point beyond which the protagonists are invested in goals. \\
TP4 - Major Setback & A critical juncture where the protagonists face significant challenges or failures. \\
TP5 - Climax & The peak of the narrative arc, encompassing the resolution of the central conflict. \\ \bottomrule
\end{tabular}
\caption{\textbf{Turning Point (TP) Types:} We describe the 5 TP types in our meso-level narrative discourse schema. A turning point is an event (or plot moment) that significantly influences a plot progression \cite{papalampidi-etal-2019-movie}. These turning points are generally in sequential order in a narrative (\textit{i.e.,} TP1 happens first; TP5 happens last).}
\label{tab:turning_points}
\vspace{-2mm}
\end{table*}

\subsection{Three Aspects of Story-Telling}

\stitle{Aspect 1: Story Arcs} A narrative's story arc charts the transformation of a story's protagonist(s) across a plot's progression. \citet{vonnegut1995shapes} developed a five-part schema to categorize story arcs. Following \citet{reagan2016emotional,wu2023word}, we adopt an expanded seven-part schema as shown in Table \ref{tab:story_arc}.
This schema captures various positive and negative transitions, such as `Rags to Riches' (i.e. a character ascends from adverse conditions to prosperity), or `Cinderella' (i.e. a character ascends from adversity, falls and then ascends again). Despite its simplicity, the story-arc classification schema has become a useful tool in writing \cite{harma2021dramatic} and computational story-telling research \cite{reagan2016emotional, chu2017ai}.  % for the simplicity with which it describes a character’s journey from the beginning to the end of the narrative.  % along an axis of 'Ill-Great Fortune'. % Not only popular among , the ontology has been widely adopted for computational storytelling research for its accessibility and its intuitive grasp of narrative dynamics . 
% These arcs serve as frameworks that guide the narrative's development, profoundly affecting story's thematic resolution.

\stitle{Aspect 2: Turning Points} A turning point in a narrative, as conceptualized by \citet{papalampidi-etal-2019-movie}, is an event (or more generally a plot-moment) that significantly influences the plot progression. Typically, a turning point represents a protagonist's transition between rises and falls %ascent or descent 
and serves to demarcate different stages of the plot.  Turning points are crucial to narratives for providing a sense of dynamism and maintaining momentum. % and audience engagement.
The types of turning points, identified by \citet{papalampidi-etal-2019-movie}, are shown in Table \ref{tab:turning_points}. Some, like ``Opportunity'', ``Change of Plans'' and ``Point of No Return'' are designed to capture exposition, or rising actions of the plot. ``Major Setback'' further develops the conflict and ``Climax'' describes the resolution. In general, we consider these last two to be the most important in determining the arc of the story.

\stitle{Aspect 3: Affective Dimensions} Two affective dimensions: arousal (\textit{i.e.}, the intensity of emotions conveyed in a sentence) and valence \textit{(i.e.}, the positivity or negativity of the emotions expressed) play crucial roles in shaping the emotional impact of narratives \cite{medhat2014sentiment}. This is quantified using the NRC-VAD lexicon \cite{mohammad-2018-obtaining}, which provides arousal and valence scores for individual tokens from a 0 to 1 scale. Affective dimensions provide a more nuanced analysis of sentence-level dynamics, capturing subtle shifts in emotional intensity and polarity that may not be fully represented in broader narrative structures, such as story arcs and turning points.

% We use affective dimensions, comprising of arousal and valence, to measure of the sentiment induced in the reader throughout the narrative. We consider affective dimensions induced both by the nature of events (i.e. which events occur) and their description (i.e. how intensely they are told). Whereas story-arc and turning point discourse markers describe structural elements shaping the narrative, arousal and valence measure the effects of the story and the art of story-telling. 
% The arousal and valence of a plot are measures of the sentiment induced in the reader throughout the narrative. We consider arousal and valence induced both by the nature of events (i.e. which events occur) and their description (i.e. how intensely they are told). Whereas story-arc and turning point discourse markers describe structural elements shaping the narrative, the arousal and valence measures more the effects of the story and the art of story-telling. 

% \yufei{TODO: talk about data collection here. 
% Proposed structure: 
% 2.1: framing
% 2.2 methods to quantify/collect data for TP, arc, and affective dimensions. }

\section{Data Collection and Annotation}\label{sec:data_collect_annotate}

Films are our culture's \textit{Gesamtkunstwerk} (or ``total work of art'') according to \citet{michelson1991your}, and our mass market vehicle for telling narratives \cite{balio2013major}. Thus, we take films as a basis for exploring the stories our culture tells itself. The narratives we examine are condensed versions of some of the most intricate storylines humans create—those found in movies. While these synopses focus on key plot developments, they should not be considered simple or straightforward.
In this section, we will describe first how we built our dataset of film plots, and then we will describe our annotation approach to study the plots' discourse structures.

\begin{figure*}[t!]
    \centering
    \includegraphics[width=0.7\linewidth]{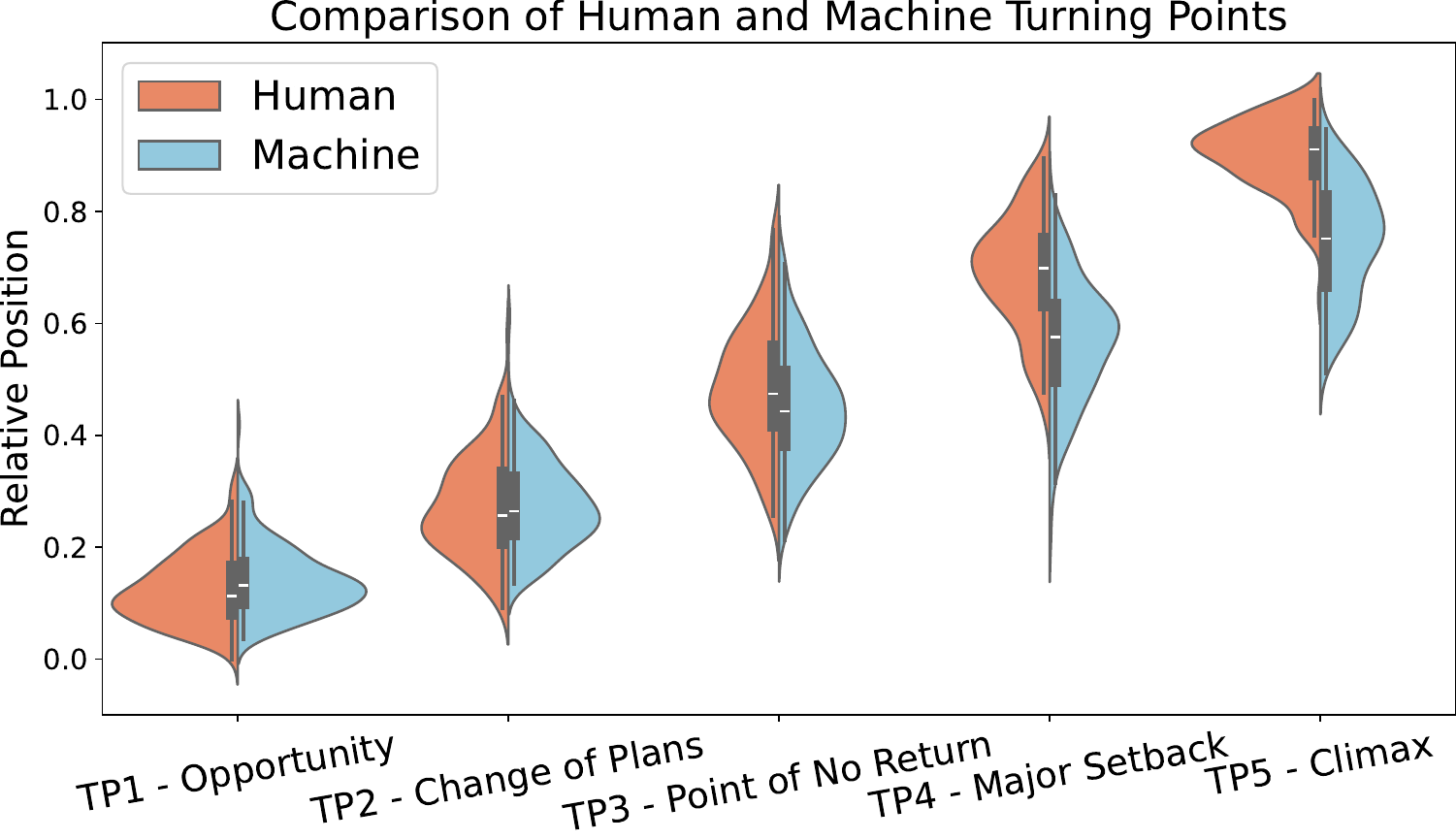}
    \vspace{-3mm}
    \caption{Violin plots showing the positions of five turning points: TP1 - opportunity, TP2 - change of plans, TP3 - point of no return, TP4 - major setback, and TP5 - climax. Relative positions (y-axis) are calculated by $\frac{\text{Index(TP$_i$)}}{\text{Total Length}}$. For example, 0.5 means that the turning point occurs exactly in the middle of the whole story. We observe early arrival for TP 4-5 in AI outputs, indicating bad pacing and a lack of intensity.}
    \vspace{-3mm}
    \label{fig:TP_positions}
\end{figure*}

\subsection{Data Preparation}

We crawl the recent English-language films category on Wikipedia\footnote{https://en.m.wikipedia.org/wiki/Category:2020s\_English-language\_films} to collect the titles, genres, release dates and synopses of these films.  %We crawl a total of 3502 films. To increase the quality of synopses, we remove those with fewer than 25 sentences, resulting in 936 movies. 
To avoid data leakage, we filter out well-known movies using the lengths of Wikipedia pages as an approximate indicator of popularity, resulting in 819 synopses. To further avoid data contamination, we rephrase the titles and initial settings by altering all the unique identifiers such as proper nouns. Finally, we instruct GPT-4 using the rephrased titles, initial settings, and the genres to generate a paired synopsis for each collected film, resulting in 1638 synopses. All human and machine narratives are roughly of the same length. 

% And then prompt GPT-4 to generate paired synopsis with the same title and initial setting. To avoid data contamination, we also rephrase the tltle and initial settings.
\subsection{Analysis Approaches}\label{subsec:data_collect}
\paragraph{Annotating Turning Point and Story Arc}

We seek to collect human annotations for each synopsis. To do so, we design annotation tasks for input narratives to %1) 
label each with a story arc, and %2) 
locate the sentential position where each of the five turning points occurs. We introduce a few key modifications to the turning point schema of \citet{papalampidi-etal-2019-movie} to handle more complex narratives: 1) flexible positioning for TP3 and TP4, and 2) optional, though discouraged, missing or multiple positions for TP2, TP3, and TP4 when annotators are uncertain. 

% This annotation task requires expertise in narrative analysis. Hence, 
We recruited 16 annotators who either hold (or are pursuing) a bachelor's degree in English or have prior experience in story analysis. To ensure the reliability of our annotators, we conducted multiple training sessions to fully onboard our annotators and administered a qualification task, exemplified in Figure~\ref{fig:survey01} in the appendix. We also designed short questions to assist annotators in determining the story arcs more accurately. %\footnote{We compile and the detailed annotation guidelines in Figures~\ref{fig:survey02}-\ref{fig:survey09} in Appendix~\ref{appendix:survey}.}
Two example pairs of human and GPT-4 written narratives, along with corresponding human annotations, are compiled in Appendix \ref{appendix:subsec:standard_annotation_example}. We also detail our annotation guidelines in Appendix \ref{appendix:survey}.

The narratives we studied have an average of 705.6 words and 37.8 sentences, longer and more complex compared to most other papers quantitatively studying narratives, such as \cite{sap2022quantifying} with 300 words and \cite{li2018annotating} with 18.5 sentences. We had a total of 440 narratives annotated, with each narrative annotated by three workers in-depth. The inter-annotator agreements (IAA) for the two tasks are measured at 0.90 (using Spearman's Correlation) and 0.62 (using Cohen's Kappa), which indicate a substantial agreement and speaks to the quality of our annotation process. 
Considering extensive labor for an in-depth human study at scale, our annotators limited their evaluations to stories produced by humans and GPT-4-0613, one of the most powerful LLMs, which should approximate the upper bound of current LLM capabilities.

\paragraph{Measuring Arousal and Valence}
We take an agentic analysis of arousal and valence, as in the previous work by \citet{field2019contextual}. 
To do this, we first instruct GPT-4 to identify the main character of the story. Then for each sentence $s_i$ in a narrative, we ask the same LLM to infer three adjectives, $W_i = \{w_{i1}, w_{i2}, w_{i3}\}$, that describe the protagonist's emotions as the plot progresses (\textit{e.g.,} amused, relaxed, anxious). We then utilize the NRC VAD lexicon \cite{mohammad-2018-obtaining} to obtain the arousal and valence scores of $w_{ij}$ ranging from 0 to 1, $j \in \{1,2,3\}$. For each sentence, we use the \textit{average} scores of $w_{ij}$ to represent the arousal and valence of $s$, obtaining $A(s_i)$ and $V(s_i)$. 
%\begin{equation*}
%    A(\mathbf{s}) = \frac{1}{|W|} \sum_{i=1}^{n} a(w_i), w_i \in W
%    \label{eqn:arousalScore}
%\end{equation*} 
%where $A(\mathbf{s})$ represents the arousal score of a sentence, and $a(w_i)$ is the arousal score of the $i^{th}$ token obtained from the NRC Word-Emotion Association lexicon
    %After obtaining individual scores for each sentence in the story, we aim to analyze how machine affective curves differ from those generated by humans. To achieve this, 
    
In our analysis, a narrative with $N$ sentences is evaluated using the discrete arousal ($A$) and valence ($S$) values at the sentence level. These values are plotted on scatter plots with sentence relative position $\frac{i}{N}$ on the x-axis, and $A(s_i)$ or $V(s_i)$ on the y-axis. To facilitate comparison across narratives of varying lengths, we interpolate these plots to generate smooth curves. The mean of these individual curves is then calculated to derive an aggregated curve that represents the arousal or valence across multiple narratives.

\section{Human vs. AI Narratives: A Discourse-Level Comparison}\label{sec:discourse_comparison}

Having described our framework for analyzing narratives, our data collection and our measurement approaches, we now describe insights we derived. %\yufei{Can move the pie chart and its corresponding paragraph to the front if everyone feels this is the #1 takeaway message}

\paragraph{\uline{LLMs incorrectly pace their storytelling relative to human writers}.}

Figure~\ref{fig:TP_positions} shows paired violin plots comparing the sentential position of turning points in human- and AI-generated stories. As shown, 
% for each turning point (i.e. TP1 (opportunity), TP2 (change of plans), TP3 (point of no return), TP4 (major setback), and TP5 (climax)) . 
while the positioning of TP1 through TP3 is consistent between human and AI narratives, we observe a substantial advancement (\textit{i.e.,} early occurrence) of TP4 and TP5 in AI outputs.  This suggests that while LLMs grasp the correct pacing to establish the initial setup (TP1, opportunity) and introduce the main goal (TP2, change of plans), they still \textbf{\textit{struggle to unfold the narrative's most crucial junctures adequately }}: major setback (TP4) and climax (TP5).
\begin{figure}[t!]
    \centering
    \includegraphics[width=0.8\linewidth]{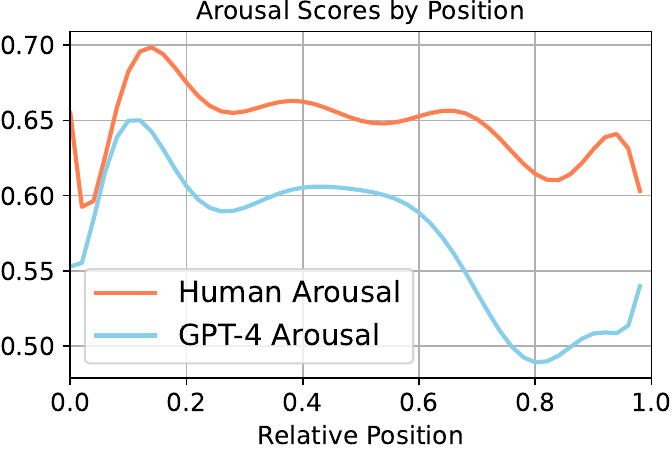}
    \vspace{-3mm}
    \caption{Arousal of human and GPT-4. Human stories consistently exhibit higher levels of suspense (greater arousal). The gap enlarges from the midpoint to the end.}
    \vspace{-1mm}
    \label{fig:arousal_curve}
\end{figure}

\paragraph{\uline{Poor pacing leads to flat narratives without suspense}.}
The pacing we observed in AI narratives, as discussed prior, is unnatural compared to human writers. It often results in less narration being spent on the last two turning points in a story (i.e. Major Setback and Climax). 
Anecdotally, we notice that when these two elements are introduced briefly and then resolved rapidly, the resulting arc feels flatter less exciting, and is more lacking in intensity. To further verify this hypothesis, we draw arousal curves in Figure~\ref{fig:arousal_curve} to visualize the suspense level throughout the whole story. We find that human-written stories consistently exhibit higher levels of suspense (\textit{i.e.}, greater arousal), but the gap begins to enlarge as the plot progresses from the midpoint (0.5 relative position) towards the end. All these observations indicate that AI-generated stories tend to be less arousing and lack suspense, especially after the introductory events are established and the action begins to build.

 \begin{figure}[t!]
    \centering
    \includegraphics[width=0.99\linewidth]{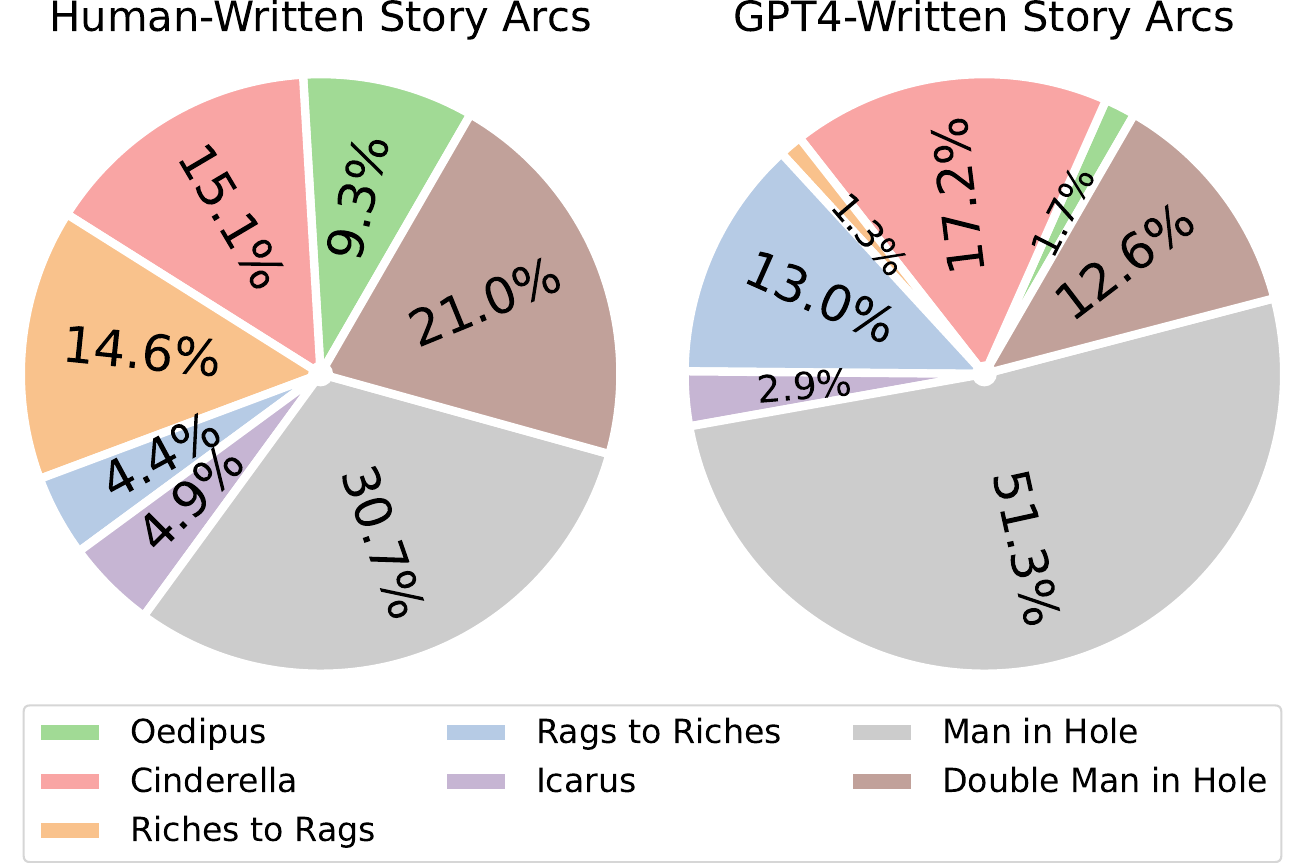}
    \vspace{-5mm}
    \caption{The share of story arcs between human and GPT-4 generated stories show significant differences. GPT-4 is much more likely to generate story arcs with less inflections and happier endings than human stories. %\yufei{too colorful? any suggestions on style?}
    }
    \vspace{-3mm}
    \label{fig:story_arc_pie_chart}
\end{figure}

\paragraph{\uline{LLMs are biased towards story arcs with positive endings and lack narrative diversity}.}
Pie charts in Figure~\ref{fig:story_arc_pie_chart} contrast the share of story arcs between human and GPT-4 generated stories. Notably, GPT-4 augments the human bias, by writing \texttt{Man in Hole}, the most popular arc type in human stories, more than half of the time.

\begin{figure}[t]
    \centering
    \includegraphics[width=0.8\linewidth]{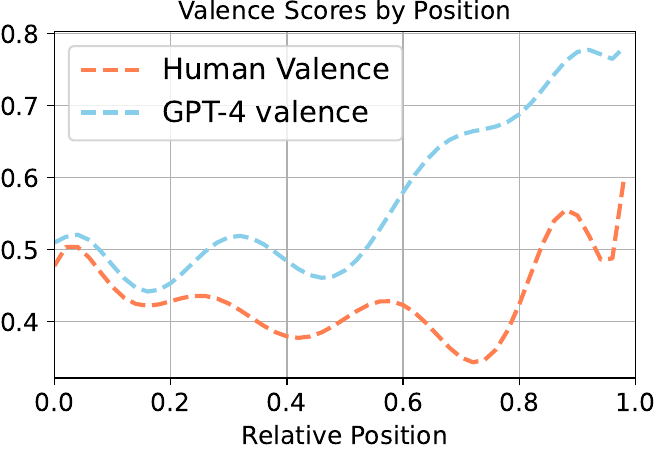}
    \vspace{-3mm}
    \caption{Valence of human and GPT-4. Human-written stories have more setbacks than GPT-4 (lower valence). The gap enlarges from the midpoint to the end.}
    \vspace{-3mm}
    \label{fig:valence_curve}
\end{figure}

Moreover, story arcs that traditionally end negatively, such as \texttt{Riches to Rags} (gradual fall) and \texttt{Oedipus} (fall then rise then fall), which represent 14.6\% and 9.3\% of human narratives, are almost missing in GPT-4 outputs (1.3\% and 1.7\%). On the other hand, \texttt{Rags to Riches} (gradual rise), which is scarcely found among human stories (4.4\%), now disproportionately accounts for 13.0\% of AI-generated stories. Such patterns lead to the conclusion that \textit{LLMs such as GPT-4 exhibit a distinct bias, strongly favoring positive outcomes and avoiding negative plot progressions.} One possible explanation is that the effect of RLHF on an LLM's language distribution pushes it more towards a positive, helpful generative stance. Figure~\ref{fig:valence_curve} also shows human-written stories contain more setbacks or negative events (less valence) while GPT-4 narratives are much more positive. Similar to arousal curves, the gap is more pronounced from the midpoint to the ending of the story.

\section{Benchmarking Narrative Comprehension}\label{sec:benchmark}
% Using the dataset collected in $\S$ \ref{subsec:data_collect}, 

We hypothesize that poor narrative comprehension of LLMs lead to their poor generative outcomes, as %wish to measure \textit{narrative reasoning} skills as a precursor to measuring \textit{generative capacities}: 
much evidence exists for these skills being tied \cite{collobert2008unified, raffel2020exploring}. Therefore, we designed and conducted two benchmark tests to measure narrative reasoning. We start by outlining our tasks ($\S$ \ref{ssec:task_formulation}) and the methodologies employed to evaluate performance ($\S$\ref{ssec:evaluation}). We finally report the benchmarking results over popular foundation LLMs ($\S$ \ref{subsec:benchmark_result}).

\subsection{Benchmark Tasks}
\label{ssec:task_formulation}
% \yufei{this sub section is too detailed. can trim. ideal length for the entire section 4: 1.25 pages.}

\paragraph{Task 1: Story Arc Identification}
Given a narrative text, our primary task is to classify it into one of several predefined story arcs. This task tests the ability of the model to understand and categorize overarching narrative structures. %The model receives as input a narrative text and is required to output a label corresponding to one of the predefined story arcs. 
The effectiveness of the model is measured by its accuracy in matching these arcs against expert annotations.

\paragraph{Task 2: Turning Point Identification}
% Given a narrative text composed of \(n\) sentences, this task focuses on the identification of five distinct turning points that critically define the trajectory of the story. Each turning point is categorized according to predefined types, which are essential for understanding the development of the narrative and its dramatic structure.
Formally, this task can be defined as follows. Given a sequence of \(n\) sentences \(S = \{s_1, s_2, \ldots, s_n\}\) that make up the narrative, the model needs to determine a set of five turning points \(T = \{t_1, t_2, t_3, t_4, t_5\}\), where each \(t_i\) is a tuple \( (p, d) \). Here, \(p\) denotes the position of the sentence within \(S\) representing the turning point, and \(d\) is a label from the predefined set of turning point types.

\paragraph{Task Variants}

We formulate two settings for each task: (1) we seek to identify turning points or story arcs given \textit{just} the text of the narrative (2) we give the model additional discourse-level features to aid in each task. Prior research has found that additional discourse information can improve narrative reasoning \cite{spangher2021multitask, spangher2024newsedits2}: we hypothesize that macro-level story discourse and meso-level information are related. More specifically, for turning point identification, \textit{information about the overarching story arc type is provided}. Conversely, when identifying story arcs, \textit{descriptions of key turning points within the narrative are included}. % This setting tests the hypothesis that additional discourse level information can aid in the models' narrative reasoning and decision-making processes.
% on the performance of language models, directly addressing the question of how discourse-level features are related.
% 
To assess how well models are able to identify story arc and turning points, we compare the model's classifications with ground truth annotations provided by human experts. %Our approach allows us to measure the degree of relation by measuring classification performance increases as a result of having additional discourse information.

\begin{table}[t]
\centering
\small
\setlength{\tabcolsep}{3pt}
\begin{tabular}{@{}lcccccc@{}}
\toprule
Model           & TP1   & TP2   & TP3   & TP4   & TP5  & Avg.  \\ \midrule

\rowcolor[HTML]{EFEFEF} Human & 59.6 & 40.3 & 37.0 & 45.4 & 50.4 & 46.6 \\\midrule
Gemini   & 40.5 & 27.3 & 15.5 & 29.4 & \textbf{43.8}  & \textbf{31.3}\\
GPT-4     & 43.9 & 20.1 & 13.8 & 23.3 & 25.4 & 25.3\\
GPT-3.5    & 28.7 & 19.5 & 8.2  & 14.9 & 23.1  & 18.8\\
Claude   & 46.5 & 24.5 & 16.3 & 30.1 & 35.7 & 30.6\\
Llama3  & 24.6 & 14.4 & 9.2  & 21.0 & 32.3 & 20.3\\

% \addlinespace % Adds a little extra space
\midrule

\textit{with arc as prior}\\

\midrule
Gemini & 38.0 & 26.1 & 12.7 & 26.1 & 40.8 & 28.7\\
GPT-4    & 38.2 & \textbf{27.6} & 13.2 & \textbf{30.3} & 27.6 &  27.3\\
GPT-3.5  & 34.6 & 16.0 & 5.1  & 11.5 & 19.9 & 17.4\\
Claude & \textbf{47.4} & 27.3 & \textbf{16.9} & 27.9 & 33.1 &  30.5\\
Llama3  & 33.5 & 15.5 & 11.0  & 20.1 & 31.0 & 22.2\\ 
\bottomrule
\end{tabular}
\caption{The success rates of five language models and humans on the task of turning point identification, presented as percentages (\%). The five turning points are TP1 - Opportunity, TP2 - Change of Plans, TP3 - Point of No Return, TP4 - Major Setback, TP5 - Climax. We use boldface to denote the best machine result. }
\label{tp_results} \vspace{-2mm}
\end{table}

\subsection{Models}
\label{ssec:evaluation}
We collect classifications from multiple state-of-the-art language models, including GPT-3.5, GPT-4, \footnote{April 29th. 2024 version}
Gemini 1.0 Pro, Claude3, Llama3-8B \cite{touvron2023Llama}. For turning point identification, we instruct a model to analyze and tag key turning points in a movie synopsis with explanations. To enhance the model's counting ability, all narratives are tagged with a sentence index. %For story arc identification, we provide task instructions along with input instance in the prompt. 
The exact prompts used are shown in Appendix \ref{apdx:prompt_details}. %\tenghao{New edit here. Please take a look.}\yufei{Give a short description of the prompt we used. Then say the detailed prompts can be found in appendix}. 
We use accuracy as the metric to measure how well the models' predictions align with expert annotations.

% can be leveraged to improve the accuracy and depth of narrative analysis performed by automated systems.

\begin{figure}[t!]
    \centering
    \includegraphics[width=0.8\linewidth]{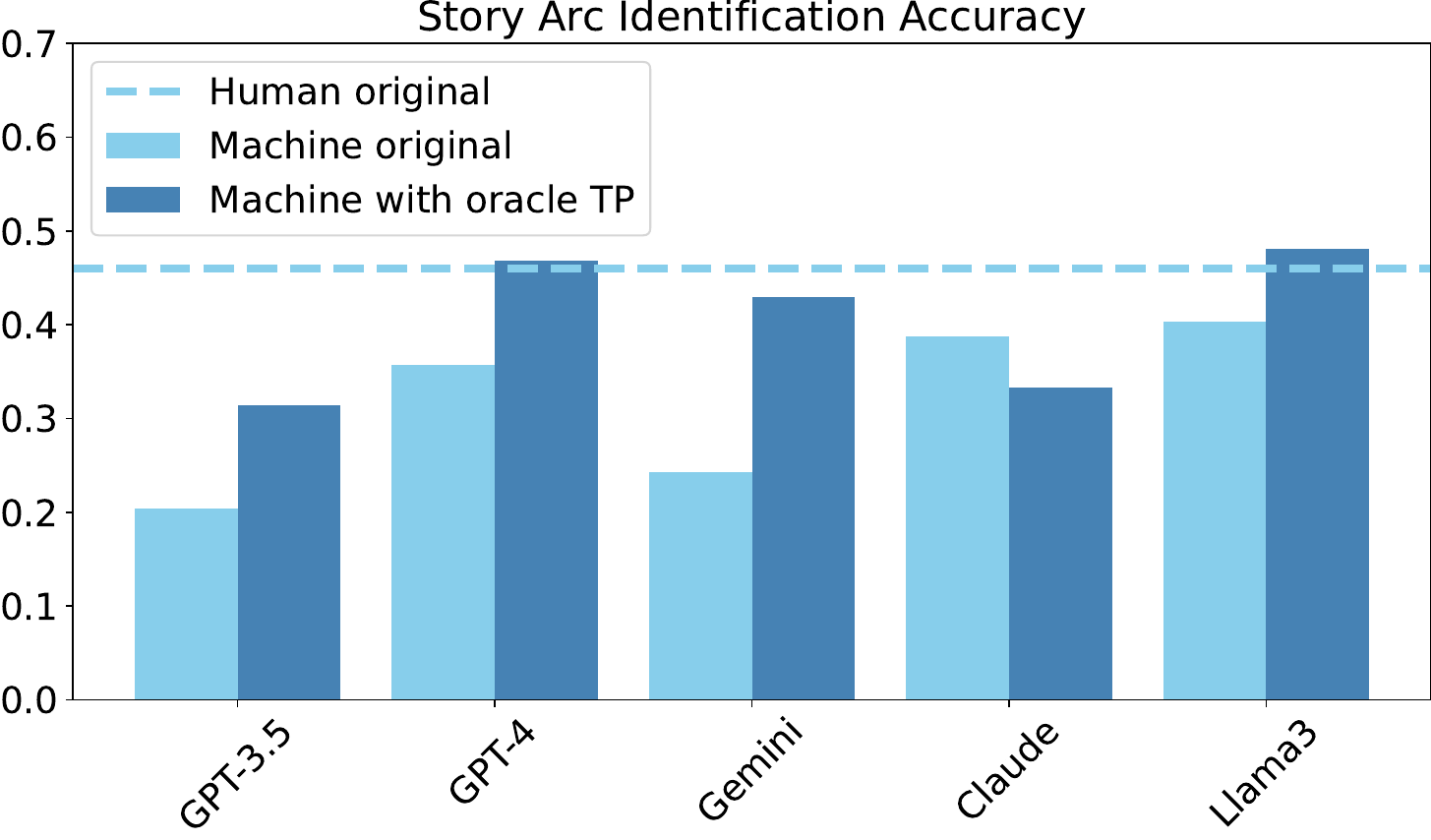}
    \vspace{-3mm}
    \caption{Story arc identification accuracy. Human judgements (blue line) are made without access to turning point information. Language models approach human accuracy \textit{only} when provided with such ground truth information, indicating the \textit{conceptual overlaps} in these discourse structures.}
    \vspace{-3mm}
    \label{fig:story_arc_exact_match}
\end{figure}

\begin{table*}[t]
\centering
\small
\setlength{\tabcolsep}{4pt}
\begin{tabular}{l ccc ccc ccc}
\toprule
 & \multicolumn{3}{c}{\textsc{Suspense}}            & \multicolumn{3}{c}{\textsc{Emotion Provoking}}             & \multicolumn{3}{c}{\textsc{Overall Preference}}             \\
 & \textbf{Best} ($\uparrow$)& \textbf{Medium} & \textbf{Worst} ($\downarrow$)& \textbf{Best} ($\uparrow$)& \textbf{Medium} & \textbf{Worst} ($\downarrow$) & \textbf{Best} ($\uparrow$)& \textbf{Medium} & \textbf{Worst} ($\downarrow$) \\ 
 \cmidrule(lr){2-4}\cmidrule(lr){5-7}\cmidrule(lr){8-10}
 
\textbf{\textit{Outline-Only}}  & 7.9\%           & 10.1\% & 82.0\%         & 14.6\%          & 24.7\% & 60.7\%          & 13.5\%          & 25.8\% & 60.7\%          \\
\textbf{\textit{+ Self-generated TP}} & \textbf{48.3\%} & 42.7\% & \textbf{9.0\%} & \uline{39.3\%}          & 42.7\% & \textbf{18.0\%} & {\uline{43.8\%}}    & 37.1\% & \textbf{19.1\%} \\
\textbf{\textit{+ Human TP}} & {\uline{46.1\%}}    & 42.7\% & {\uline{11.2\%} }  & \textbf{48.3\%} & 28.1\% & {\uline{ 23.6\%}}    & \textbf{44.9\%} & 32.6\% & {\uline{22.5\%}}  \\  \bottomrule
\end{tabular}
\caption{Human evaluated results in suspense, emotion provocation, and overall preference. We compare machine generations with and without the awareness of turning points (TP3, TP4, and TP5). % We compare three machine generation approaches: outline-only, + self-generated TPs, and + human-given TPs.
}
\label{tab:tp4gen}
\end{table*}

\subsection{Benchmark Findings}
\label{subsec:benchmark_result}

\paragraph{Larger, closed models identify turning point identification with higher accuracy}
Table~\ref{tp_results} reports the model performance on the turning point identification task without incorporating story arc information as a prior. Gemini and Claude demonstrate the highest performance, with average accuracies over 30\%, respectively. GPT-4 also performs reasonably well with an average accuracy of 26\%. However, GPT-3.5 and Llama3 lag behind. All models perform below human levels, emphasizing the challenge of accurately identifying story arcs using current LLMs.
%, with average accuracies of 18.8\% and 20.3\%, respectively. 
%These results indicate that larger and more sophisticated models are better equipped to handle the complexities of narrative structures.

\paragraph{Story arc identification also lags human performance}
Figure~\ref{fig:story_arc_exact_match} shows each model's performance on identifying story arc types. The original model performances (light blue bars) reveal that accuracy is generally low across all models. For instance, GPT-3.5 achieves an exact agreement score of approximately 0.2 (random guessing being $\frac{1}{7}=0.14$). GPT-4, %reaches around 0.35. 
Claude, and Llama3 perform better, with exact agreements above 0.35. Similar to turning point, human still achieves higher accuracy than the LLMs without additional knowledge.
%Llama3-8B shows the best performance among the tested models with a score approaching 0.5. 

\paragraph{Incorporating additional discourse information improves model comprehension}
We find that finer-grained information will benefit the coarse-grained task more than the reverse.
For example, when the ground-truth, macro-level discourse tag (\textit{i.e.,} story arc) is provided to the meso-level task (\textit{i.e.,} turning point identification), the average accuracy of a few LLMs (GPT-4 and Llama3) improves by 2\%. However, not all models benefit from such hints. On the other hand, the dark blue bars in Figure~\ref{fig:story_arc_exact_match} demonstrate a significant improved performance in story arc identification when turning point information is given across all models. %GPT-4's score increases from 0.35 to about 0.46, and Gemini shows an increase from 0.3 to around 0.4. Llama3-8B, which already had the highest original performance, improves further to a score above 0.5. 
Both results support our hypothesis that \emph{incorporating discourse-level features can enhance the machine's narrative reasoning capabilities}.

%This suggests that giving additional narrative context can enhance their ability to predict turning points with greater accuracy.

\section{Towards Better Machine Storytelling}
\label{sec:enhance}
Finally, we investigate whether incorporating the discourse aspects into the generation stage enhances machine's storytelling ability.

\noindent \textbf{Reasoning about TPs improves overall narrative construction}, including \uline{reduced plot holes} and \uline{enhanced suspense and emotion provocation.} Motivated by our observations that a major flaw in vanilla LLM story-telling is narrative pacing (in $\S$\ref{sec:discourse_comparison}) we hypothesize that integrating discourse features can improve pacing and significantly improve narratives.

We test three variations of a planning-first \cite{yao2019plan} approach (i.e. generating first the outline and then the narrative). Each variation incorporates different degrees of explicit structure.

\begin{itemize}[topsep=2pt, itemsep=-2pt, leftmargin=*]
    \item \textit{Outline-Only}: We simply instruct the model to generate an outline, then expand it to a full story.
    \item \textit{+ Self TPs}: We instruct the LLM to generate an outline that marks the 3rd, 4th, and last turning points (i.e. ``Point of No Return'', ``Major Setback'', and ``Climax''), with their detailed definitions, and then to write the full-length story.
    \item \textit{+ Human TPs}: We replace the machine-generated major setback and climax with the oracle, human-crafted equivalents (which are typically more compelling and intriguing than their machine-generated counterparts).
\end{itemize} % Following the work of \cite{yang2022re3}, we implement a planning-based narrative generation framework that initially produces a story outline. 
% This outline is subsequently expanded into a full-length narrative , a process we refer to as the \textit{original} story generation approach.

% \citet{chakrabarty2023creativity} proposes a novel human-in-the-loop collaborative writing paradigm, prompting us to quantitatively investigate the impact of substituting machine-generated turning points with those authored by humans. 
We annotate comparatively, ranking three randomly shuffled narratives in terms of suspense, emotion provocation, and overall preference. Table \ref{tab:tp4gen} shows win-rates over the above three approaches. Both \textit{+ Self TP} and \textit{+Human TP} achieve significant gains over \textit{Outline-Only}, highlighting the efficacy of incorporating TPs in LLM-generated narratives.
% Interestingly, we find that incorporating the events of oracle turning points doesn't lead to higher quality stories.
Interestingly, we find that while \textit{+Human TP} scored highly, especially for emotional engagement (48.3\%), it is not significantly preferred over \textit{+ Self TP}. Upon further investigation, we realize that the enforcement of external events in \textit{+Human TP} could disrupt the machine's narrative flow, making the whole plot illogical at times. \textit{+ Self TP}, which maintained the natural flow of LLM with its own generations, emerged as the most balanced and least disliked approach. This indicates that future work in the domain of human-machine collaborative writing must be careful to integrate human creativity in beneficial ways.  % \yufei{suggest to not directly use Variant1/2/3 as names.}

%While Variant 3 being a coarse intergration of human-machine collaborative writing, future research could look into this and propose better collaboration frameworks.

\paragraph{Incorporating explicit directives about story arcs helps improve narrative diversity}

\begin{table}[!t]
\small
\centering
\begin{tabular}{lclc}
\toprule
\textbf{Requested Arc} & \textbf{Acc.} & \textbf{Requested Arc} & \textbf{Acc.} \\
\cmidrule(lr){1-2}\cmidrule(lr){3-4}
Cinderella          & 33\% & Oedipus             & 64\% \\
Riches to Rags      & 33\% & Icarus              & 67\% \\
Double Man in Hole  & 54\% & Man in Hole         & 71\% \\
Rags to Riches      & 57\% & Average             & 54\% \\

\bottomrule
\end{tabular}
\caption{GPT-4 shows poor accuracy in generating narratives with specified story arc types, although is better for arcs that have one inflection point (e.g. ``Man in the Hole'') compared with two (e.g. ``Cinderella'').}
\label{tab:requested_arc_success} \vspace{-2mm}
\end{table}
\begin{table}[!t]
\centering
\small
\setlength{\tabcolsep}{2pt}
\begin{tabular}{@{}lccccc@{}}
\toprule
      Diversity& {\scriptsize\textsc{Theme}} & {\scriptsize\textsc{Setting}} & {\scriptsize\textsc{Conflict}} & {\scriptsize\textsc{Character}} & {\scriptsize\textsc{Overall}} \\ \midrule
\textbf{\textit{Outline-Only}} & 5\%  & 32\%          & 5\%  & 23\% & 23\% \\
\textbf{\textit{Tie}}    & 32\% & \textbf{36\%} & 41\% & 27\% & 9\%  \\
\textbf{\textit{Arc-Enhanced}} & \textbf{64\%}   & 32\%             & \textbf{55\%}     & \textbf{50\%}      & \textbf{68\%}   \\ \bottomrule
\end{tabular}
\caption{Win rates of the outline-only stories and story-arc enhanced stories. We focus on four specific aspects of diversity: theme, setting, conflict, and character.}\vspace{-2mm}
\label{tab:arc_diversity}
\end{table}

Motivated by our observations, in figure \ref{fig:story_arc_pie_chart}, that LLM generations lack arc-level diversity, we explore whether explicit instruction can induce a more human-like story arcs. We design another variant,  \textit{Arc Enhanced}, that explicitly instructs the model to generate story with a specified story arc, specifies the number of major rises and falls and details the initial and ending state of the protagonists. 

First, we evaluate how well LLMs are able to follow the requested story-arcs. The results, in Table~\ref{tab:requested_arc_success}, show that GPT-4 achieves an average success rate below 55\%, suggesting that LLMs' capability to mirror human narrative distributions is limited, even with explicit instructions. Notably, these models struggle with story arcs that depict negative plot progressions (\textit{e.g.,} riches to rags), humble starts (\textit{e.g.,} cinderella), and those with complex narrative dynamics (\textit{e.g.,} double man in hole). 

Next, we compare the narrative diversity between sets of \textit{Arc Enhanced} stories and \textit{Outline-Only} stories\footnote{We instruct annotators to examine diversity in the following aspects: 1) \textbf{Thematic}: the central ideas conveyed; 2) \textbf{Setting}: when and where these stories take place; 3) \textbf{Conflict type}: including but not limited to character with self, other characters, society, nature, technology, fate; 4) \textbf{Character}: including but not limited to personality, background, development, and relationships}. As seen in Table \ref{tab:arc_diversity}, \textit{Arc Enhanced} significantly outperforms \textit{Outline-Only} across most aspects of diversity that we considered: thematic, conflict-type, and character. We conclude that story arc discourse is a significant driver of many aspects of narrative diversity, affirming the basic truth of \citet{vonnegut1995shapes}'s assertion that stories can be broadly categorized into story arc types.%\footnote{, which we report in Appendix \ref{appendix:subsec:feedback}.}
%

%, In terms of settings, both approaches were comparable, each receiving 32\%. For conflict type, the ``Arc Enhanced'' approach was favored by 55\% of annotators, compared to 5\% for the original. Character diversity was also more prominent with the ``Arc Enhanced'' approach, with 50\% of annotators noting greater diversity compared to 23\% for the original. Overall, the ``Arc Enhanced'' approach was preferred by 68\% of annotators, while the original received 23\%. the ``Arc Enhanced'' approach significantly improves narrative diversity, especially in themes, conflict types, and character development, enhancing the richness and variety of the generated stories. \tenghao{Maybe some powerful takeaway messages here}.

\section{Qualitative Study: Understanding Human Preferences}

\begin{figure}[!t]
    \centering
    \includegraphics[width=1\linewidth]{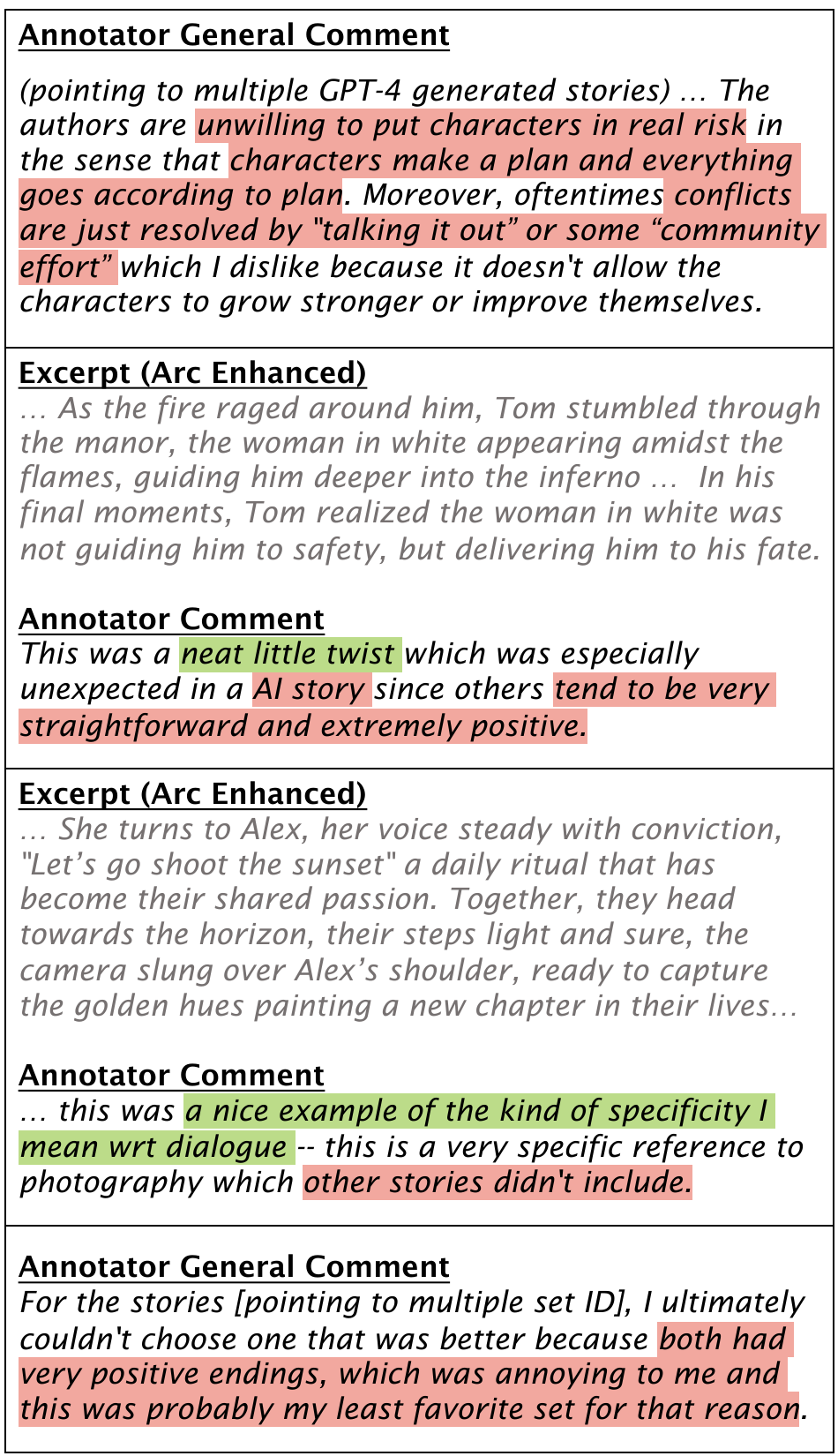}
    \vspace{-2mm}
    \caption{Human annotators' feedback on machine-generated stories when comparing the \textit{Outline-Only} vs \textit{Arc-Enhanced} strategy. They were blind to the prompting strategies and all presented stories were randomly shuffled. We reconstruct their comments and color-code with green for favorable ones and red for unfavorable ones. Continued in Figure \ref{fig:interview_2}.}
    \label{fig:interview_1}
    \vspace{-1em}
\end{figure}

After completing all annotation tasks, we conducted two follow-up interviews to gather qualitative feedback from our annotators. These interviews focused on annotators who worked on the task in $\S$ \ref{sec:enhance} --- reading pairs of machine generated narratives (\textit{Outline-Only} vs \textit{Arc-Enhanced}) and examining diversities plus overall preference. They were encouraged to freely provide justifications or comments on any of the readings. After the interviews, we reconstructed their feedback and presented representative comments in Figure \ref{fig:interview_1} and Figure \ref{fig:interview_2}. \uline{Overall, the human annotators prefer concrete narratives with twists in plot development that are logical and well-motivated. They dislike straightforward, positive plots or those with `miracle turns' that are not adequately justified.}

\section{Related Work}

\stitle{Discourse-Aware Evaluation} In contrast to conventional story evaluation frameworks, which primarily focus on fluency and coherence, \textit{discourse-aware} evaluation focuses on critiquing the structural and creative quality of machine-generated content \cite{harel2024measuring, spangher2024llm_planning,tian2024detecting}. \citet{liu2024unlocking} introduced a model that assesses stories by embedding conventional narrative structures within the evaluation process. Complementing this, \citet{chakrabarty2024art} explore narrative differences between humans and AI through a qualitative study and \cite{li2024predicting} reveal that RLHF-aligned language models are less diverse than the base LMs..
\citet{begus2023experimental} delves into the creative outputs of LLMs, questioning their true creativity versus their capacity to merely replicate observed patterns, which encourages further exploration of the creative limits of these models. Additionally, \citet{wang2023learning} introduces the Positional Discourse Coherence metric to quantitatively assess logical narrative progression. 

However, prior works have been limited by the vague definitions of creativity and discourse structure. We take inspiration from the literature theories of discourse analysis of drama and fiction \cite{labov1967narrative, vonnegut1995shapes}. For instance, \citet{freytag1894technik} proposed a
a five-part dramatic structure, now commonly understood as Exposition, Rising Action, Climax, Falling Action, and Dénouement. \citet{li2018annotating} combined such literature theories and annotated story macro-structures. In non-fiction storytelling like news writing, \citet{choubey2020discourse, spangher2021multitask, spangher2022sequentially} demonstrated that language models can classify similar elements, with artificial news differing from human-generated content. Likewise, we define specific story arcs and turning points in creative stories and examine how stories generated by LMs structurally differ from human ones.

\stitle{Discourse-Aware Generation with LLMs} %Story generation has been a popular direction. Early works mainly focus on generating coherent and logical stories \cite{yao2019plan,yang2022re3}. 
Attempts to incorporate discourse features into story generation include \citet{yao2019plan,han2022go,yang2022re3} that focus on generating coherent, logical, and interesting stories. \citet{huang-etal-2023-affective} incorporates affective dimensions to foster the creation of more captivating stories. \citet{brei2024returning} examines the efficacy of ``bookends'' as a structural enhancement in narrative quality. Further studies have also ventured into embedding elements of suspense to forge more engaging narratives \cite{Zehe2023TowardsAC, Xie2024CreatingSS}. Different from previous endeavors, our study enhances narrative construction through the systematic incorporation of discourse elements, similar to \citet{spangher2022sequentially} who focus on news structures. Our approach seeks to bridge the gap between human-like storytelling and the capabilities of current AI systems through three levels of discourse elements. 

\section{Conclusion}
This work aims to advance the understanding and generation of narratives through the lens of three discourse elements: story arcs at macro-level, turning points at meso-level, and affective dimensions at micro-level. We contribute an expert-annotated dataset, based on which we conducted quantitative comparison between human and AI in terms of narrative generation and comprehension: LLMs fall short especially in story writing. We find models lack narrative diversity, and struggle at develop crucial turning points, such as major setback and climax, leading to less engaging stories.
We also show promising results that discourse-aware generation improves AI's story-telling ability in terms of suspense, emotion engagement, and narrative diversity. 

We view our effort as a useful starting point towards a systematic analysis of narrative discourse. We hope the collected dataset and experimental results, along with our proposed perspective, will attract wider academic interest in discourse studies and provide insights into better narrative generation, comprehension, and evaluation.

\section*{Limitations}

For both discourse-level comparison ($\S$ \ref{sec:data_collect_annotate}) and better machine storytelling  ($\S$ \ref{sec:enhance}) which require in-depth human annotation, we limit our experiment to GPT-4 generated narratives. While we believe the conclusions are applicable to other LLMs such as Claude, Llama, Gemini, etc., their generations are not direct assessed. 

Another limitation is that our research primarily focuses on English-based LLMs and resources. Our initial focus on English allows us to establish the discourse-level framework before expanding to others. Future research can look into expanding this scope to include multilingual language models and diverse linguistic resources. This expansion could help to better understand and predict flavors across different cultural and linguistic contexts, potentially uncovering unique insights and flavor combinations that are specific to various cuisines and regional preferences. 

Earlier studies, such as \citet{Huang2021UncoveringIG}, have identified the presence of gender biases within this dataset. Consequently, we want to point out that generating stories using our pipeline may also be at risk of perpetuating and intensifying these biases.

%Another limitation of our study lies in the 
% \yufei{acl checklist also asks us to decribe the potential risks}

\section*{Acknowledgement}
The authors would like to thank the PlusLab members at UCLA and the anonymous reviewers for their valuable feedback and helpful discussions. 
This research is partly supported by National Science Foundation CAREER award \#2339766, an Amazon AGI foundation research award, and a Google Research Scholar grant. 
Tenghao Huang and Jonathan May are supported by the Defense Advanced Research Projects Agency with award HR00112220046.
\newpage
\bibliography{anthology,custom}

\begin{thebibliography}{49}
\expandafter\ifx\csname natexlab\endcsname\relax\def\natexlab#1{#1}\fi

\bibitem[{AI@Meta(2024)}]{llama3modelcard}
AI@Meta. 2024.
\newblock \href {https://github.com/meta-llama/llama3/blob/main/MODEL_CARD.md} {Llama 3 model card}.

\bibitem[{Anthropic(2024)}]{anthropic2024claude}
AI~Anthropic. 2024.
\newblock The claude 3 model family: Opus, sonnet, haiku.
\newblock \emph{Claude-3 Model Card}, 1.

\bibitem[{Balio(2013)}]{balio2013major}
Tino Balio. 2013.
\newblock A major presence in all of the world's important markets: The globalization of hollywood in the 1990s.
\newblock In \emph{Contemporary Hollywood Cinema}, pages 58--73. Routledge.

\bibitem[{Begus(2023)}]{begus2023experimental}
Nina Begus. 2023.
\newblock Experimental narratives: A comparison of human crowdsourced storytelling and ai storytelling.
\newblock \emph{arXiv preprint arXiv:2310.12902}.

\bibitem[{Brei et~al.(2024)Brei, Zhao, and Chaturvedi}]{brei2024returning}
Anneliese Brei, Chao Zhao, and Snigdha Chaturvedi. 2024.
\newblock \href {https://doi.org/10.18653/v1/2024.naacl-short.10} {Returning to the start: Generating narratives with related endpoints}.
\newblock In \emph{Proceedings of the 2024 Conference of the North American Chapter of the Association for Computational Linguistics: Human Language Technologies (Volume 2: Short Papers)}, pages 101--112, Mexico City, Mexico. Association for Computational Linguistics.

\bibitem[{Chakrabarty et~al.(2024)Chakrabarty, Laban, Agarwal, Muresan, and Wu}]{chakrabarty2024art}
Tuhin Chakrabarty, Philippe Laban, Divyansh Agarwal, Smaranda Muresan, and Chien-Sheng Wu. 2024.
\newblock Art or artifice? large language models and the false promise of creativity.
\newblock In \emph{Proceedings of the CHI Conference on Human Factors in Computing Systems}, pages 1--34.

\bibitem[{Choubey et~al.(2020)Choubey, Lee, Huang, and Wang}]{choubey2020discourse}
Prafulla~Kumar Choubey, Aaron Lee, Ruihong Huang, and Lu~Wang. 2020.
\newblock Discourse as a function of event: Profiling discourse structure in news articles around the main event.
\newblock In \emph{Proceedings of the 58th Annual Meeting of the Association for Computational Linguistics}.

\bibitem[{Chu et~al.(2017)Chu, Dunn, Roy, Sands, and Stevens}]{chu2017ai}
Eric Chu, Jonathan Dunn, Deb Roy, Geoffrey Sands, and Russell Stevens. 2017.
\newblock Ai in storytelling: Machines as cocreators.
\newblock \emph{McKinsey \& Company Media \& Entertainment}.

\bibitem[{Collobert and Weston(2008)}]{collobert2008unified}
Ronan Collobert and Jason Weston. 2008.
\newblock A unified architecture for natural language processing: Deep neural networks with multitask learning.
\newblock In \emph{Proceedings of the 25th international conference on Machine learning}, pages 160--167.

\bibitem[{Field et~al.(2019)Field, Bhat, and Tsvetkov}]{field2019contextual}
Anjalie Field, Gayatri Bhat, and Yulia Tsvetkov. 2019.
\newblock Contextual affective analysis: A case study of people portrayals in online\# metoo stories.
\newblock In \emph{Proceedings of the international AAAI conference on web and social media}, volume~13, pages 158--169.

\bibitem[{Freytag(1894)}]{freytag1894technik}
Gustav Freytag. 1894.
\newblock \emph{Die technik des dramas}.
\newblock S. Hirzel.

\bibitem[{Han et~al.(2022)Han, Chen, Tian, and Peng}]{han2022go}
Rujun Han, Hong Chen, Yufei Tian, and Nanyun Peng. 2022.
\newblock Go back in time: Generating flashbacks in stories with event temporal prompts.
\newblock In \emph{Proceedings of the 2022 Conference of the North American Chapter of the Association for Computational Linguistics: Human Language Technologies}, pages 1450--1470.

\bibitem[{Harel-Canada et~al.(2024)Harel-Canada, Zhou, Muppalla, Yildiz, Kim, Sahai, and Peng}]{harel2024measuring}
Fabrice~Y Harel-Canada, Hanyu Zhou, Sreya Muppalla, Zeynep~Senahan Yildiz, Miryung Kim, Amit Sahai, and Nanyun Peng. 2024.
\newblock Measuring psychological depth in language models.
\newblock In \emph{Proceedings of The 2024 Conference on Empirical Methods in Natural Language Processing (EMNLP)}.

\bibitem[{H{\"a}rm{\"a} et~al.(2021)H{\"a}rm{\"a}, K{\"a}rkk{\"a}inen, and Jeronen}]{harma2021dramatic}
Kimmo H{\"a}rm{\"a}, Sirpa K{\"a}rkk{\"a}inen, and Eila Jeronen. 2021.
\newblock The dramatic arc in the development of argumentation skills of upper secondary school students in geography education.
\newblock \emph{Education Sciences}, 11(11):734.

\bibitem[{Hu et~al.(2021)Hu, Liu, Thomsen, Gao, and Nielbo}]{hu2021dynamic}
Qiyue Hu, Bin Liu, Mads~Rosendahl Thomsen, Jianbo Gao, and Kristoffer~L Nielbo. 2021.
\newblock Dynamic evolution of sentiments in never let me go: Insights from multifractal theory and its implications for literary analysis.
\newblock \emph{Digital Scholarship in the Humanities}, 36(2):322--332.

\bibitem[{Huang et~al.(2021)Huang, Brahman, Shwartz, and Chaturvedi}]{Huang2021UncoveringIG}
Tenghao Huang, Faeze Brahman, Vered Shwartz, and Snigdha Chaturvedi. 2021.
\newblock \href {https://api.semanticscholar.org/CorpusID:237502736} {Uncovering implicit gender bias in narratives through commonsense inference}.
\newblock \emph{ArXiv}, abs/2109.06437.

\bibitem[{Huang et~al.(2023)Huang, Qasemi, Li, Wang, Brahman, Chen, and Chaturvedi}]{huang-etal-2023-affective}
Tenghao Huang, Ehsan Qasemi, Bangzheng Li, He~Wang, Faeze Brahman, Muhao Chen, and Snigdha Chaturvedi. 2023.
\newblock \href {https://doi.org/10.18653/v1/2023.findings-emnlp.789} {Affective and dynamic beam search for story generation}.
\newblock In \emph{Findings of the Association for Computational Linguistics: EMNLP 2023}, pages 11792--11806, Singapore. Association for Computational Linguistics.

\bibitem[{Kaniss(1991)}]{kaniss1991making}
Phyllis Kaniss. 1991.
\newblock \emph{Making local news}.
\newblock University of Chicago Press.

\bibitem[{Kasneci et~al.(2023)Kasneci, Se{\ss}ler, K{\"u}chemann, Bannert, Dementieva, Fischer, Gasser, Groh, G{\"u}nnemann, H{\"u}llermeier et~al.}]{kasneci2023chatgpt}
Enkelejda Kasneci, Kathrin Se{\ss}ler, Stefan K{\"u}chemann, Maria Bannert, Daryna Dementieva, Frank Fischer, Urs Gasser, Georg Groh, Stephan G{\"u}nnemann, Eyke H{\"u}llermeier, et~al. 2023.
\newblock Chatgpt for good? on opportunities and challenges of large language models for education.
\newblock \emph{Learning and individual differences}, 103:102274.

\bibitem[{Labov and Waletzky(1967)}]{labov1967narrative}
William Labov and Joshua Waletzky. 1967.
\newblock Narrative analysis.
\newblock In \emph{Essays on the Verbal and Visual Arts}. University of Washington Press, Seattle, WA.

\bibitem[{Langer(1942)}]{langer1942philosophy}
Susanne~K Langer. 1942.
\newblock \emph{Philosophy in a new key: A study in the symbolism of reason, rite, and art}.
\newblock Harvard University Press.

\bibitem[{Li et~al.(2018)Li, Cardier, Wang, and Metze}]{li2018annotating}
Boyang Li, Beth Cardier, Tong Wang, and Florian Metze. 2018.
\newblock Annotating high-level structures of short stories and personal anecdotes.
\newblock In \emph{Proceedings of the Eleventh International Conference on Language Resources and Evaluation (LREC 2018)}.

\bibitem[{Li et~al.(2024)Li, Shi, Pagnoni, West, and Holtzman}]{li2024predicting}
Margaret Li, Weijia Shi, Artidoro Pagnoni, Peter West, and Ari Holtzman. 2024.
\newblock Predicting vs. acting: A trade-off between world modeling \& agent modeling.
\newblock \emph{arXiv preprint arXiv:2407.02446}.

\bibitem[{Liu et~al.(2024)Liu, Su, Shareghi, and Collier}]{liu2024unlocking}
Yinhong Liu, Yixuan Su, Ehsan Shareghi, and Nigel Collier. 2024.
\newblock Unlocking structure measuring: Introducing pdd, an automatic metric for positional discourse coherence.
\newblock \emph{arXiv preprint arXiv:2402.10175}.

\bibitem[{Medhat et~al.(2014)Medhat, Hassan, and Korashy}]{medhat2014sentiment}
Walaa Medhat, Ahmed Hassan, and Hoda Korashy. 2014.
\newblock Sentiment analysis algorithms and applications: A survey.
\newblock \emph{Ain Shams engineering journal}, 5(4):1093--1113.

\bibitem[{Michelson(1991)}]{michelson1991your}
Annette Michelson. 1991.
\newblock " where is your rupture?": Mass culture and the gesamtkunstwerk.
\newblock \emph{October}, 56:43--63.

\bibitem[{Minaee et~al.(2024)Minaee, Mikolov, Nikzad, Chenaghlu, Socher, Amatriain, and Gao}]{minaee2024large}
Shervin Minaee, Tomas Mikolov, Narjes Nikzad, Meysam Chenaghlu, Richard Socher, Xavier Amatriain, and Jianfeng Gao. 2024.
\newblock Large language models: A survey.
\newblock \emph{arXiv preprint arXiv:2402.06196}.

\bibitem[{Mohammad(2018)}]{mohammad-2018-obtaining}
Saif Mohammad. 2018.
\newblock \href {https://doi.org/10.18653/v1/P18-1017} {Obtaining reliable human ratings of valence, arousal, and dominance for 20,000 {E}nglish words}.
\newblock In \emph{Proceedings of the 56th Annual Meeting of the Association for Computational Linguistics (Volume 1: Long Papers)}, pages 174--184, Melbourne, Australia. Association for Computational Linguistics.

\bibitem[{OpenAI(2022)}]{chatgpt2022}
OpenAI. 2022.
\newblock \href {https://openai.com/blog/chatgpt} {Introducing chatgpt}.

\bibitem[{OpenAI(2023)}]{openai2023gpt4}
OpenAI. 2023.
\newblock \href {http://arxiv.org/abs/2303.08774} {Gpt-4 technical report}.

\bibitem[{Papalampidi et~al.(2019)Papalampidi, Keller, and Lapata}]{papalampidi-etal-2019-movie}
Pinelopi Papalampidi, Frank Keller, and Mirella Lapata. 2019.
\newblock \href {https://doi.org/10.18653/v1/D19-1180} {Movie plot analysis via turning point identification}.
\newblock In \emph{Proceedings of the 2019 Conference on Empirical Methods in Natural Language Processing and the 9th International Joint Conference on Natural Language Processing (EMNLP-IJCNLP)}, pages 1707--1717, Hong Kong, China. Association for Computational Linguistics.

\bibitem[{Raffel et~al.(2020)Raffel, Shazeer, Roberts, Lee, Narang, Matena, Zhou, Li, and Liu}]{raffel2020exploring}
Colin Raffel, Noam Shazeer, Adam Roberts, Katherine Lee, Sharan Narang, Michael Matena, Yanqi Zhou, Wei Li, and Peter~J Liu. 2020.
\newblock Exploring the limits of transfer learning with a unified text-to-text transformer.
\newblock \emph{Journal of machine learning research}, 21(140):1--67.

\bibitem[{Reagan et~al.(2016)Reagan, Mitchell, Kiley, Danforth, and Dodds}]{reagan2016emotional}
Andrew~J Reagan, Lewis Mitchell, Dilan Kiley, Christopher~M Danforth, and Peter~Sheridan Dodds. 2016.
\newblock The emotional arcs of stories are dominated by six basic shapes.
\newblock \emph{EPJ data science}, 5(1):1--12.

\bibitem[{Reid et~al.(2024)Reid, Savinov, Teplyashin, Lepikhin, Lillicrap, Alayrac, Soricut, Lazaridou, Firat, Schrittwieser et~al.}]{reid2024gemini}
Machel Reid, Nikolay Savinov, Denis Teplyashin, Dmitry Lepikhin, Timothy Lillicrap, Jean-baptiste Alayrac, Radu Soricut, Angeliki Lazaridou, Orhan Firat, Julian Schrittwieser, et~al. 2024.
\newblock Gemini 1.5: Unlocking multimodal understanding across millions of tokens of context.
\newblock \emph{arXiv preprint arXiv:2403.05530}.

\bibitem[{Sap et~al.(2022)Sap, Jafarpour, Choi, Smith, Pennebaker, and Horvitz}]{sap2022quantifying}
Maarten Sap, Anna Jafarpour, Yejin Choi, Noah~A Smith, James~W Pennebaker, and Eric Horvitz. 2022.
\newblock Quantifying the narrative flow of imagined versus autobiographical stories.
\newblock \emph{Proceedings of the National Academy of Sciences}, 119(45):e2211715119.

\bibitem[{Spangher et~al.(2024{\natexlab{a}})Spangher, Huang, Cho, and May}]{spangher2024newsedits2}
Alexander Spangher, Kung-Hsiang Huang, Hyundong Cho, and Jonathan May. 2024{\natexlab{a}}.
\newblock Newsedits 2.0: Learning the intentions behind updating news.

\bibitem[{Spangher et~al.(2021)Spangher, May, Shiang, and Deng}]{spangher2021multitask}
Alexander Spangher, Jonathan May, Sz-Rung Shiang, and Lingjia Deng. 2021.
\newblock Multitask semi-supervised learning for class-imbalanced discourse classification.
\newblock In \emph{Proceedings of the 2021 conference on empirical methods in natural language processing}, pages 498--517.

\bibitem[{Spangher et~al.(2022)Spangher, Ming, Hua, and Peng}]{spangher2022sequentially}
Alexander Spangher, Yao Ming, Xinyu Hua, and Nanyun Peng. 2022.
\newblock Sequentially controlled text generation.
\newblock In \emph{Findings of the Association for Computational Linguistics: EMNLP 2022}, pages 6848--6866.

\bibitem[{Spangher et~al.(2024{\natexlab{b}})Spangher, Peng, Gehrmann, and Dredze}]{spangher2024llm_planning}
Alexander Spangher, Nanyun Peng, Sebastian Gehrmann, and Mark Dredze. 2024{\natexlab{b}}.
\newblock Do llms plan like human writers? comparing journalist coverage of press releases with llms.
\newblock In \emph{Proceedings of The 2024 Conference on Empirical Methods in Natural Language Processing (EMNLP)}.

\bibitem[{Tian et~al.(2024)Tian, Pan, and Peng}]{tian2024detecting}
Yufei Tian, Zeyu Pan, and Nanyun Peng. 2024.
\newblock Detecting machine-generated long-form content with latent-space variables.
\newblock \emph{Findings of The 2024 Conference on Empirical Methods in Natural Language Processing (EMNLP)}.

\bibitem[{Touvron et~al.(2023)Touvron, Martin, Stone, Albert, Almahairi, Babaei, Bashlykov, Batra, Bhargava, Bhosale et~al.}]{touvron2023Llama}
Hugo Touvron, Louis Martin, Kevin Stone, Peter Albert, Amjad Almahairi, Yasmine Babaei, Nikolay Bashlykov, Soumya Batra, Prajjwal Bhargava, Shruti Bhosale, et~al. 2023.
\newblock Llama 2: Open foundation and fine-tuned chat models.
\newblock \emph{arXiv preprint arXiv:2307.09288}.

\bibitem[{Van~Dijk(1980)}]{van1980macrostructures}
Teun~A Van~Dijk. 1980.
\newblock \emph{Macrostructures: An interdisciplinary study of global structures in discourse, interaction, and cognition}.
\newblock Lawrence Erlbaum.

\bibitem[{Vonnegut(1995)}]{vonnegut1995shapes}
Kurt Vonnegut. 1995.
\newblock \href {https://www.youtube.com/watch?v=oP3c1h8v2ZQ} {Shapes of stories}.
\newblock Accessed June 2024.

\bibitem[{Wang et~al.(2023)Wang, Yang, Zhu, Yang, Cohen, Li, and Tian}]{wang2023learning}
Danqing Wang, Kevin Yang, Hanlin Zhu, Xiaomeng Yang, Andrew Cohen, Lei Li, and Yuandong Tian. 2023.
\newblock Learning personalized story evaluation.
\newblock \emph{arXiv preprint arXiv:2310.03304}.

\bibitem[{Wu et~al.(2023)Wu, Wang, and Mihalcea}]{wu2023word}
Winston Wu, Lu~Wang, and Rada Mihalcea. 2023.
\newblock Word category arcs in literature across languages and genres.
\newblock In \emph{Proceedings of the The 5th Workshop on Narrative Understanding}, pages 36--47.

\bibitem[{Xie and Riedl(2024)}]{Xie2024CreatingSS}
Kaige Xie and Mark Riedl. 2024.
\newblock \href {https://api.semanticscholar.org/CorpusID:268032393} {Creating suspenseful stories: Iterative planning with large language models}.
\newblock In \emph{Conference of the European Chapter of the Association for Computational Linguistics}.

\bibitem[{Yang et~al.(2022)Yang, Tian, Peng, and Klein}]{yang2022re3}
Kevin Yang, Yuandong Tian, Nanyun Peng, and Dan Klein. 2022.
\newblock Re3: Generating longer stories with recursive reprompting and revision.
\newblock In \emph{Proceedings of the 2022 Conference on Empirical Methods in Natural Language Processing}, pages 4393--4479.

\bibitem[{Yao et~al.(2019)Yao, Peng, Weischedel, Knight, Zhao, and Yan}]{yao2019plan}
Lili Yao, Nanyun Peng, Ralph Weischedel, Kevin Knight, Dongyan Zhao, and Rui Yan. 2019.
\newblock Plan-and-write: Towards better automatic storytelling.
\newblock In \emph{Proceedings of the AAAI Conference on Artificial Intelligence}, volume~33, pages 7378--7385.

\bibitem[{Zehe et~al.(2023)Zehe, Schr{\"o}ter, and Hotho}]{Zehe2023TowardsAC}
Albin Zehe, Juliane Schr{\"o}ter, and Andreas Hotho. 2023.
\newblock \href {https://api.semanticscholar.org/CorpusID:258615734} {Towards a computational analysis of suspense: Detecting dangerous situations}.
\newblock \emph{ArXiv}, abs/2305.06818.

\end{thebibliography}
\newpage
\appendix

\clearpage
\begin{figure}[t!]
    \centering
    \includegraphics[width=0.9\linewidth]{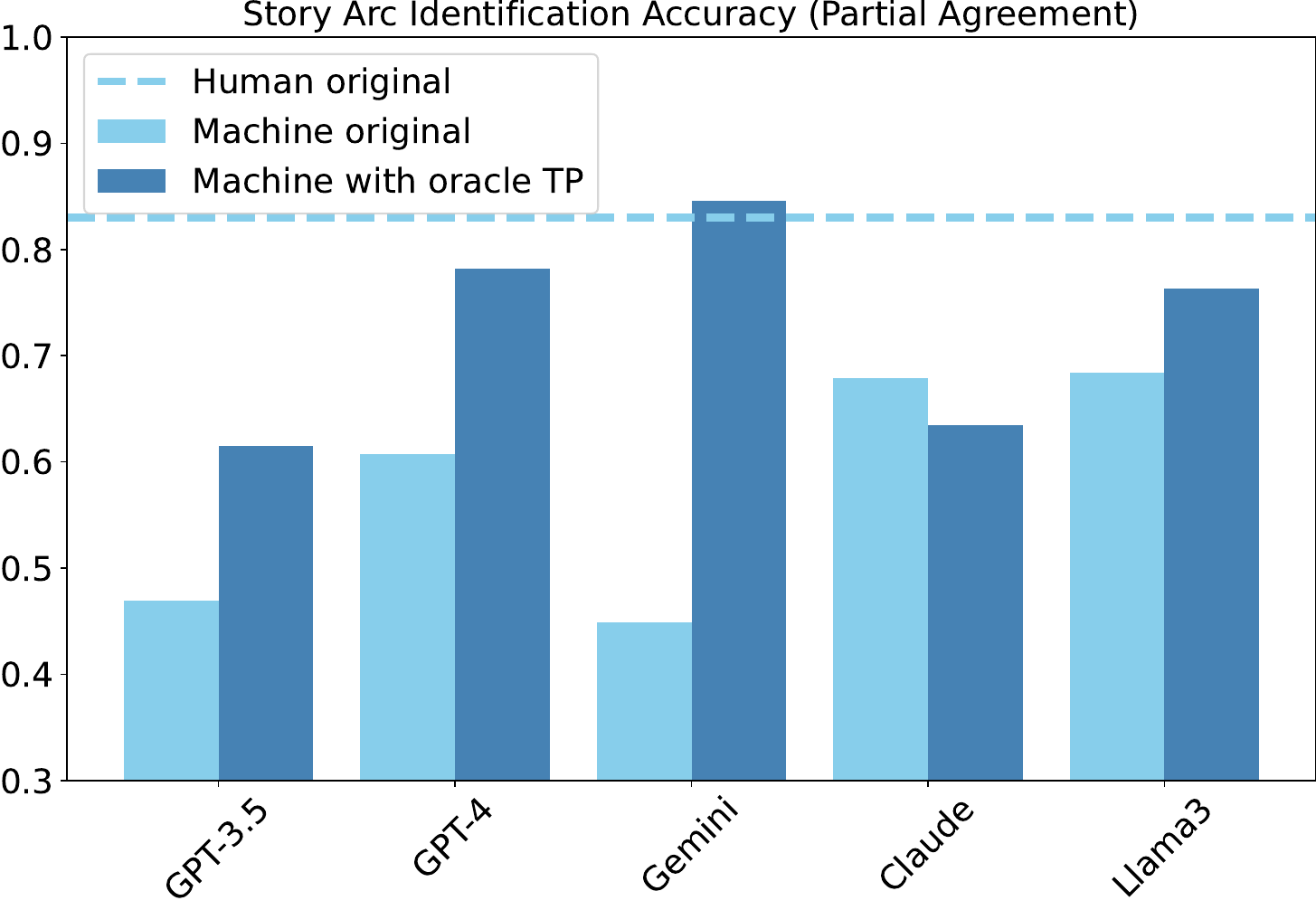}
    \vspace{-3mm}
    \caption{Story arc fuzzy matching results.}
    \vspace{-3mm}
    \label{fig:story_arc_partial_match}
\end{figure}
\label{partial_results}

\begin{table*}[h!]
\centering
\small
\begin{tabular}{c|c}
\toprule
\textbf{Story Arc} & \textbf{Hard Label Pairs} \\
\midrule
Man in Hole & Double Man in Hole, Cinderella \\
%\hline
Double Man in Hole & Man in Hole, Cinderella \\
%\hline
Cinderella & Rags to Riches, Man in Hole, Double Man in Hole \\
%\hline
Rags to Riches & Cinderella \\
%\hline
Riches to Rags & Oedipus \\
%\hline
Oedipus & Riches to Rags \\
\bottomrule
\end{tabular}
\caption{Hard label pairs for story arcs.}
\label{hard_label_pairs}
\end{table*}

\begin{table}[t]
\centering
\small
\setlength{\tabcolsep}{3pt}
\begin{tabular}{@{}lccccc@{}}
\toprule
Model           & TP1   & TP2   & TP3   & TP4   & TP5  \\ \midrule

\rowcolor[HTML]{EFEFEF} Human &88.2 & 72.3 & 68.9 & 77.3 & 82.4 \\\midrule
GEMINI   & 89.2 & 65.5 & 60.8 & \textbf{65.5} & \textbf{83.0} \\
GPT4     & 82.5 & 54.5 & 50.8 & 58.2 & 71.4 \\
GPT35    & 78.5 & 57.4 & 40.5 & 51.8 & 74.4 \\
CLAUDE   & 87.2 & 64.8 & 51.0 & 60.7 & 79.6 \\
LLAMA3  & 73.8 & 46.7 & 43.6 & 59.5 & 79.5 \\

% \addlinespace % Adds a little extra space
\midrule

\textit{with arc as prior}\\

\midrule
GEMINI & 85.2 & 69.0 & 53.5 & 62.7 & 82.4 \\
GPT4    & 78.9 & \textbf{78.9} & \textbf{57.9} & 60.5 & 76.3 \\
GPT35  & 78.2 & 52.6 & 32.7 & 44.9 & 76.3 \\
CLAUDE & \textbf{87.7} & 68.8 & 55.8 & 62.3 & 77.3 \\
LLAMA3  & - & - & - & - & - \\ 
\bottomrule
\end{tabular}
\caption{The fuzzy matching success rates of five language models and humans on the task of turning point identification, presented as percentages (\%).}
\label{tp_results_fuzzy} \vspace{-2mm}
\end{table}

\section{Additional Results}
Due to the level of abstraction from detailed stories to arcs, we recognize that even human annotators experience discrepancies when categorizing a story into a specific story arc type \cite{hu2021dynamic}. For example, what one person might dismiss as a moderate obstacle that will \textit{not} affect the overall arc could be perceived by another as more significant, leading to variations in the number of falls assigned to the chosen story arc. To address this, we identified story arcs that by nature blur together or are easily confused, which we call ``hard pairs'' as illustrated in Table \ref{hard_label_pairs}. We then employed fuzzy matching to assess the models' capability in identifying story arcs, by granting credit to a model if its predicted label falls within the hard pairs, even if it is not an exact match with the ground truth. Figure \ref{hard_label_pairs} presents the results of the fuzzy matching on story arc identification. Under this setting, all models perform significantly below human levels, highlighting the inherent challenges in accurately identifying story arcs using LLMs. It is worth noting that the performance gap is \textbf{larger} than story arc exact match results, as shown in Figure \ref{fig:story_arc_exact_match}.

We also apply a similar fuzzy matching approach to the task of turning point identification. Instead of crediting the model exactly selecting the groudtruth sentence for a turning point, we now credit a model for choosing a sentence that is within $\pm 3$ positions of the groundtruth sentence. We report turning points fuzzy matching results in Figure \ref{tp_results_fuzzy}. We observe that the overall trend is not significantly different compared to the results from exact matches.

\section{Examples of Generated Narratives and Annotations}\label{appendix:example_narratives}

\begin{figure}[!t]
    \centering
    \includegraphics[width=1\linewidth]{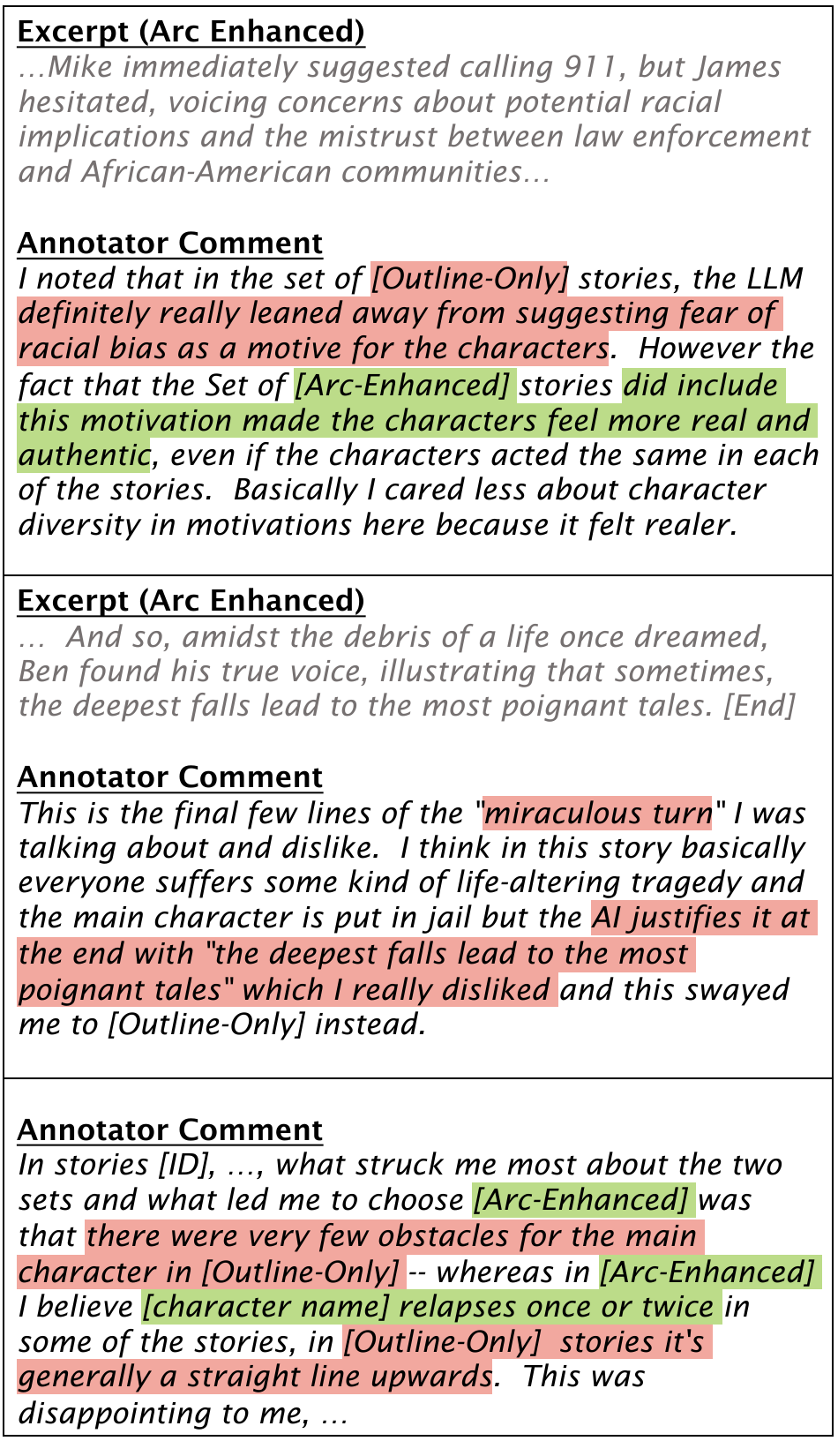}
    \caption{Figure \ref{fig:interview_1} Continued. Prior to the interviews and after the all annotation tasks were completed, annotators were already informed that the presented stories were generated by AI, but the specific methods and models were hidden from them. }
    \label{fig:interview_2}
\end{figure}

\subsection{Examples of standard annotation} \label{appendix:subsec:standard_annotation_example}
We exemplify two pairs of human and machine written narratives along with their marked annotations in Table \ref{example:human_movie1} to Table \ref{example:machine_movie2}. In both cases, human-written narratives have more suspenseful and arousing events; the major setbacks and climax arrive earlier in machine generated narratives.

\subsection{Detailed feedback from interviewed annotators} \label{appendix:subsec:feedback}

We are lucky to have conducted interviews and collected detailed feedback from two annotators who worked for the last task--- reading pairs of machine generated narratives (\textit{Outline-Only} vs \textit{Arc-Enhanced}) and examining diversities plus overall preference ($\S$ \ref{sec:enhance}). Before the interview, annotators were told that all the presented stories were generated by AI, but with different methods which were hidden from them. They were asked to freely provide any justification or comment on any of the readings. We reconstruct their comments and report representative ones in Figure \ref{fig:interview_1} and Figure \ref{fig:interview_2}. \uline{Overall, the human annotators prefer concrete narratives with twists in plot development that are logical and well-motivated. They dislike straightforward, positive plots or those with `miracle turns' but are not adequately justified.}

\begin{table*}[t!]
\small
\caption{Example 1 of human written narratives and the annotated story arc, turning points.}\label{example:human_movie1}
\begin{tabularx}{\textwidth}{p{1cm}|X}
\toprule
\multicolumn{2}{c}{\textbf{Source}: Human} \\
\multicolumn{2}{c}{\textbf{Title}: The Dark and the Wicked}\\
\multicolumn{2}{c}{\textbf{Genre}: Horror}\\
\multicolumn{2}{c}{\textbf{Annotated Story Arc}:  Riches to Rags}     \\ \midrule
                                                                                                                        
1         & Siblings Louise and Michael return to their family farm in Texas when their father's chronic illness seems to be reaching its last stages.     \\
2         & Their mother seems disturbed at their arrival, and expresses a desire for the children to leave.                                               \\
\rowcolor[HTML]{C6E0B4} 3 (tp1)        & That night, she hangs herself in the barn after (apparently involuntarily) cutting off her own fingers in the kitchen.                         \\
4         & As time goes on, Louise and Michael start to understand what happened to their mother.                                                         \\
5 &
  Their father’s nurse confides in them that she heard their mother whispering to their father, but it seemed as if she was speaking not to him, but some other presence. \\
\rowcolor[HTML]{C6E0B4} 6  (tp2)       & Michael finds their mother's diary, which describes her fears of an unnamed and possibly demonic presence preying on her husband.              \\
7         & At their mother's burial, Louise and Michael meet Father Thorne, a priest who claims to have known their mother.                               \\
8         & Later that night, Father Thorne appears at the farm, beckoning them from outside, before vanishing before their eyes.                          \\
9 &
  Meanwhile, Charlie, a ranch hand who lives on a nearby plot of land in his RV, witnesses a vision of what appears to be Louise, speaking indistinctly and cutting herself repeatedly with a kitchen knife. \\
10        & The entity drives a distraught Charlie to shoot himself in the head with his shotgun.                                                          \\
11        & Louise is subsequently unable to reach Charlie by phone, unaware that he is dead.                                                              \\
12        & Louise calls the phone number that Father Thorne gave her to ask why he visited the farm the night prior.                                      \\
13        & The man who answers claims to have never met her, and says that he lives in Chicago and has never been to Texas.                               \\
\rowcolor[HTML]{C6E0B4}  14 (tp3)       & Worried for their father's safety, the siblings summon a doctor for a house call and request that he be moved to a hospital.                   \\
15        & The doctor determines that their father's health is grave, and that he is on his deathbed.                                                     \\
16        & He tells the siblings he cannot relocate him to a hospital, as moving him could result in him dying en route.                                  \\
17        & On the farm, Louise and Michael find that their large herd of goats have all been brutally killed.                                             \\
18        & The two start a bonfire to dispose of the numerous animal carcasses.                                                                           \\
19        & That night, Michael is approached in the barn by an apparition of his nude mother, who disappears as she approaches him.                       \\
20 &
  Later, while Louise lies in bed beside her father, she has a nightmare in which the entity attempts to possess her, but she manages to resist it, before witnessing her father levitating against the ceiling. \\
21        & In the morning, Charlie's granddaughter arrives at the farm and informs Louise that he killed himself two days prior.                          \\
22        & The girl's forlorn demeanor soon turns malevolent, and Louise realizes it is in fact the entity taking the shape of Charlie's granddaughter.   \\
23        & She too disappears before Louise's eyes.                                                                                                       \\
24        & The nurse arrives moments later to care for Louise and Michael's father.                                                                       \\
\rowcolor[HTML]{C6E0B4} 25  (tp4)      & Meanwhile, Louise finds that Michael has fled the farm to return to his wife and daughters, leaving her behind.                                \\
26        & Michael calls Louise from his cell phone, and tells her she too should leave.                                                                  \\
27 &
  Moments later, the nurse, possessed by the entity, begins stabbing herself with a pair of knitting needles, attacking Louise in the process, before stabbing herself in the eyes, killing herself. \\
\rowcolor[HTML]{C6E0B4} 28  (tp5)      & Michael arrives at his home and finds the dead bodies of his daughters and wife in the kitchen, in what appears to have been a murder–suicide. \\
29        & A distraught Michael cuts his own throat, and moments after, observes that the bodies of his wife and daughters have disappeared.              \\
30        & He realizes the entity has tricked him when his wife and daughters enter the house moments later, and find him bleeding to death.              \\
31        & Back at the farm, Louise regains consciousness at nightfall and finds that her father is dying.                                                \\
32        & Moments after he dies, the demon in turn attacks and claims her.                                                                               \\ \bottomrule
\end{tabularx}
\end{table*}

\begin{table*}[]
\small
\caption{Example 1 of machine written narratives and the annotated story arc, turning points.}\label{example:machine_movie1}
\begin{tabularx}{\textwidth}{p{1cm}|X}
\toprule
\multicolumn{2}{c}{\textbf{Source}: Machine} \\
\multicolumn{2}{c}{\textbf{Title}: The Dark and the Wicked (slightly rephrased to avoid memorization)}\\
\multicolumn{2}{c}{\textbf{Genre}: Horror}\\
\multicolumn{2}{c}{\textbf{Annotated Story Arc}:  Man in Hole}     \\ \midrule
1 &
  The lives of siblings Amy and John Mackenzie, leading peaceful lives amid the ups and downs of New York City, put a vibrant appeal to the audience. \\
2 &
  As the story unfolds, Amy, a well-renowned interior designer, graces her office with an artistic flair. \\
3 &
  At the same time, John, engrossed in complex coding, generates buzzing innovative software for a top-tier tech firm. \\
\rowcolor[HTML]{C6E0B4}4 (tp1)&
  All's well in their city life until an unanticipated phone call from a family lawyer pierces their harmonious routine like a sharp edged blade. \\
5 &
  The news delivered shakes their peaceful existence, revealing their estranged father's terminal pancreatic cancer diagnosis. \\
6 &
  In response to the heart-wrenching news, they decide to leave their hustle-bustle life centered in New York and retreat to their tranquil origins of Monterey, California. \\
7 &
  The transition from the city to the contemplative countryside period calls them to assist their ailing father. \\
8 &
  Greeted by a frail figure that once used to be their lively father, they begin to comprehend the severity of the situation. \\
9 &
  The ancestral ranch, once verdant with lush vineyards, had transformed into a deserted expanse with signs of virtually no upkeep. \\
\rowcolor[HTML]{C6E0B4} 10 (tp2) &
  Amid all the despair, the siblings start noticing a series of weird incidents that were out of sync with the placid ranch. \\
11 &
  Favorite books getting relocated from their accustomed spots, the sudden rush of cold gust, and cellar doors creaking open during the silent night hours become regular events. \\
12 &
  The duo, fueled by heightened curiosity, takes up the challenge to unfold the source of these unnatural occurrences. \\
13 &
  Their quest leads them to an overshadowed wooden cabin hidden under years of vegetation, shrouded in dust and sprinkled with ancestral memories. \\
\rowcolor[HTML]{C6E0B4} 14 (tp3) &
  The cabin unveils a century-old diary hidden beneath a rotting carpet, narrating tales of their ancestors intertwined with hints of supernatural events. \\
15 &
  The diary begins painting centuries-old tales of unsettling shadows and persistent cold gusts that have made the ranch their abode- striking a chilling accord with the incidents happening on the ranch currently. \\
16 &
  A cryptic message on a diary page sends shivers down their spine and reveals a terrifying secret about their bloodline curse. \\
\rowcolor[HTML]{C6E0B4} 17 (tp4)&
  Days turn into nights and the paranormal activities begin escalating, correlating with their father's alarming health detoriation. \\
18 &
  The haw-haw of their father rapidly morphs into weak choking gasps for breath. \\
19 &
  Seeking guidance from the compassionate local Sacred Heart Church priest, the siblings begin decoding the family curse etched within the old diary. \\
20 &
  Consumed by fear and a worryingly less amount of time, they connect the dots that point towards their ancestors' grave mistake- the theft of land from Native Americans. \\
21  &
  As they make amends for their ancestors' misdeeds, they notice a significant reduction in the paranormal activities around the ranch. \\
 \rowcolor[HTML]{C6E0B4} 22 (tp5)&
 More importantly, their father's gasps for air and trembling substantially reduce, and his breaths start stabilizing. \\
23 &
  Transformations take over the ranch. \\
24 &
  Nights become serene, and paranormal events decline, allowing peace to bask onto the property. \\
25 &
  The siblings are warmly embraced as brave warriors by the locals for enduring and persevering through the haunting incidents and lifting the ancestral curse from their family. \\
26 &
  The ranch starts buzzing with life as locals lend their hands in restoring the property. \\
27 &
  Laughs echo through the vast expanse, replacing the eerie events that had earlier dominated the place. \\
28 &
  The bustling New York City life that was earlier their home becomes eventually replaced with vivid memories of open landscapes and the haunting experience at the ranch. \\
29 &
  The twilight, once a symbol of dread and fear, now stands as a testament of their bravery and acquired wisdom. \\
30 &
  The movie concludes with a panoramic view of the ranch, brimming with life against the beautiful sunset. \\
31 &
  The Mackenzies' ranch, once residing under the shadow of a curse, had finally seen a new dawn. \\ \bottomrule
\end{tabularx}
\end{table*}

\begin{table*}[]
\small
\caption{Example 2 of human written narratives and the annotated story arc, turning points.}\label{example:human_movie2}
\begin{tabularx}{\textwidth}{p{1cm}|X}
\toprule
\multicolumn{2}{c}{\textbf{Source}: Human} \\
\multicolumn{2}{c}{\textbf{Title}: Fair Play}\\
\multicolumn{2}{c}{\textbf{Genre}: Thriller}\\
\multicolumn{2}{c}{\textbf{Annotated Story Arc}:  Riches to Rags}     \\ \midrule
1 &
  Emily Meyers and Luke Edmunds, analysts at the cutthroat Manhattan hedge fund One Crest Capital, are in a secret passionate relationship unbeknownst to their coworkers. \\
2 &
  Luke proposes to Emily while at his brother's wedding, and she happily accepts. \\
3 &
  The next day, one of the company's portfolio managers is fired. \\
4 &
  Emily tells Luke she overheard her colleagues mentioning Luke being considered as a replacement and they celebrate that night. \\
 \rowcolor[HTML]{C6E0B4} 5 (tp1) &
  However, at a late-night meeting with Campbell, the firm's CEO, Emily learns she will be receiving the promotion. \\
6 &
  Emily reluctantly breaks the news to Luke, but he expresses his support. \\
 \rowcolor[HTML]{C6E0B4} 7 (tp2)&
  As Emily settles into her new job, Luke's resentment over not being promoted becomes increasingly apparent, leading to tensions in his relationship with Emily. \\
8 &
  Luke becomes consumed with the work of a self-help guru coaching people on how to assert themselves in the workplace. \\
9 &
  When Emily questions his choice to spend \$3,000 on the course, Luke suggests she could benefit from becoming more assertive, to which she becomes defensive. \\
10 &
  Luke rebuffs Emily's attempts to initiate sex and goes to bed. \\
11 &
  While out for drinks with Campbell and Paul, a senior executive at the fund, Emily learns Campbell is seeking to get rid of Luke, considering him ineffectual. \\
12 &
  Emily attempts to advocate more for Luke in the workplace, but it backfires when Luke makes a poor trading call that loses the company \$25 million, leading to Campbell insulting her. \\
13 &
  Luke attempts to rectify himself by feeding Emily insider information confirming the alleged collapse of a company whose stock the fund can short. \\
14 &
  Concerned about the trade being illegal, Emily recommends Campbell to short another company, which proves successful. \\
15 &
  When the short sale is closed, Emily receives a \$575,000 commission check. \\
 \rowcolor[HTML]{C6E0B4} 16 (tp3)&
  Emily considers celebrating her success with Luke, who is in her office after hours to discuss strategies for future trades but opts to go to a strip club with her male co-workers. \\
17 &
  She comes home intoxicated while Luke, after seeing the check, has no interest in having sex with her. \\
18 &
  When another portfolio manager is fired the next day, Luke wants Emily to recommend him for the role, but she hints Campbell is not interested in promoting him. \\
19 &
  Luke goes to Campbell's office and makes an elaborate speech pledging his loyalty to him, only to learn Campbell has already hired a new portfolio manager. \\
20 &
  That night, Emily learns her mother had planned a surprise engagement party for them that Friday. \\
21 &
  A drunken Luke accuses Emily of stealing his job, but Emily reveals Campbell wanted to fire him, leading Luke to storm out. \\
 \rowcolor[HTML]{C6E0B4} 22 (tp4)&
  The next day, while Emily, Campbell, and Paul pitch to overseas investors, Luke barges into the conference room intoxicated and causes a scene, berating Campbell for denying him a promotion and revealing his relationship with Emily, which has violated company policy since her promotion. \\
23 &
  An infuriated Emily is unable to reach Luke over the phone, only to find him at the engagement party. \\
24 &
  The two argue in front of their families, and Emily smashes a bottle on Luke's head when he suggests she had traded sexual favors for the promotion. \\
25 &
  Emily retreats to a bathroom where Luke finds her and the two argue before having sex. \\
26 &
  During sex, Luke forces Emily forward twice, causing her face to slam against the bathroom counter. \\
27 &
  Emily tells Luke to stop, but he does not. \\
28 &
  The next morning, to protect her job, Emily tells Campbell she was being stalked by Luke and they were never in a relationship. \\
29 &
  Emily returns home to find Luke there, having packed up his belongings and planning to move in with his brother. \\
 \rowcolor[HTML]{C6E0B4} 30 (tp5)&
  Infuriated by his nonchalant attitude and demanding an apology for raping her, Emily threatens Luke with a knife. \\
31 &
  She attacks Luke with the knife until he apologizes and breaks down crying. \\
32 &
  Luke begs for her forgiveness and Emily orders him to leave before dropping the knife and smiling. \\ \bottomrule
\end{tabularx}
\end{table*}

\begin{table*}[]
\small
\caption{Example 2 of machine written narratives and the annotated story arc, turning points.}\label{example:machine_movie2}
\begin{tabularx}{\textwidth}{p{1cm}|X}
\toprule
\multicolumn{2}{c}{\textbf{Source}: Machine} \\
\multicolumn{2}{c}{\textbf{Title}: Fair Play (slightly rephrased to avoid memorization)}\\
\multicolumn{2}{c}{\textbf{Genre}: Thriller}\\
\multicolumn{2}{c}{\textbf{Annotated Story Arc}:  Man in Hole}     \\ \midrule
1 &
  The story unfolds with a picturesque view of Wall Street's towering skyscrapers, where our protagonists, Madison Carter and Noah Mitchell, are immersed in complex financial reports. \\
2  & The scene transitions to display Madison and Noah at the heart of Two Peak Enterprises.                            \\
3  & Their relationship, albeit business-oriented, is a vital component of the firm.                                    \\
4  & Despite the demands of their high-pressure jobs, the two manage to cultivate a budding romance.                    \\
 \rowcolor[HTML]{C6E0B4}5 (tp1) & Suddenly, a transaction of unprecedented proportions starts to shake the foundations of Two Peaks.                 \\
6  & This billion-dollar deal begins to dominate the conversation, raising stakes across the financial landscape.       \\
7  & Madison and Noah's covert relationship comes under scrutiny.                                                       \\
8  & There's a simultaneous tension between their fear of exposure and the stress of the impending billion-dollar deal. \\
 \rowcolor[HTML]{C6E0B4} 9 (tp2) & In their analysis of the gargantuan deal, Madison and Noah identify a discrepancy that raises alarm bells.         \\
10 &
  They launch a secret investigation, delving furiously into the labyrinth of deception that infects the higher echelons of Two Peaks. \\
11 &
  Their once-familiar office morphs into unwelcoming territory as unknown adversaries send chilling threats in response to their probing. \\
12 &
  Their interpersonal bonds are tested further as they wrestle with a moral quandary surrounding their loyalty to their employer and their duty as conscientious citizens. \\
 \rowcolor[HTML]{C6E0B4} 13 (tp3) & A new roadblock appears; an unexpected internal audit is announced at Two Peaks.                                   \\
14 & Madison and Noah escalate their investigation.                                                                     \\
 \rowcolor[HTML]{C6E0B4} 15 (tp4) &
  Their actions lead to an earth-shattering revelation: the very person they respected and emulated, their boss, is the puppeteer behind the corruption. \\
16 & Upon the discovery of their boss's puppeteering, retaliation is swift and vicious.                                 \\
17 & A gripping chase ensues with Madison and Noah racing against time to secure irrefutable evidence of the crimes.    \\
18 & They manage to secure the valuable evidence necessary to expose their nefarious boss.                              \\
19 & They take a moral stand, choosing their ethical responsibility over their coveted careers.                         \\
 \rowcolor[HTML]{C6E0B4} 20 (tp5)&
  Armed with damning evidence, Madison and Noah confront their colleagues at Two Peaks' annual gala, unveiling their CEO's unscrupulous activities to an astounded audience. \\
21 & Amid the chaos, a physical altercation erupts.                                                                     \\
22 & Madison and Noah courageously battle the onslaught to protect their evidence.                                      \\
23 & After the ordeal, they hand over their evidence to the authorities.                                                \\
24 & The downfall of Two Peak Enterprises sends shockwaves across Wall Street.                                          \\
25 & Madison and Noah, hailed as righteous heroes, decide to distance themselves from the aggressive world of finance.  \\
26 & The narrative closes with the couple embarking on a new life in a bucolic setting.                                 \\
27 & A note of suspense strikes as hints point at an omnipresent surveillance.                                          \\
28 & The screen pans to a computer monitor, with Two Peaks' now-defunct website displayed.                              \\
29 & The narrative ends leaving a lasting sense of suspense.                             \\ \bottomrule
\end{tabularx}
\end{table*}

\section{Experimental Details}

\subsection{Human Annotation Interfaces}\label{appendix:survey}
\begin{figure*}[t!]
    \centering
    \includegraphics[width=0.95\linewidth]{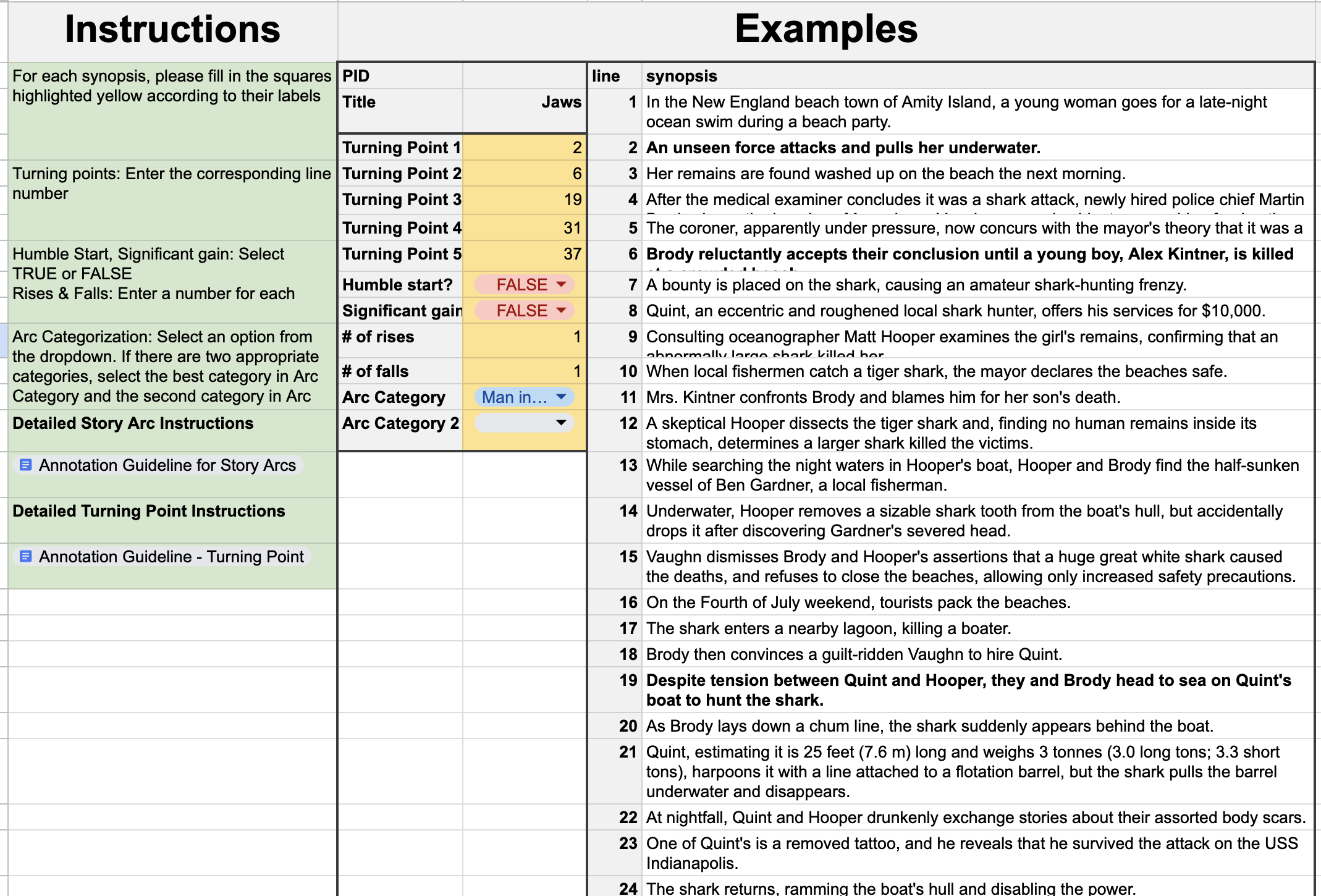}
    \vspace{-2mm}
    \caption{Human Annotation Interface for Turning Point and Story Arc.
}
    \label{fig:survey01}
    \vspace{-5mm}
\end{figure*}

Recall that in $\S$ \ref{sec:data_collect_annotate} we design annotation tasks for input narratives to 1) 
label each with a story arc, and 2) 
locate the sentential position where each of the five turning points occurs. An example of the task interface can be found in Figure \ref{fig:survey01}.

\begin{figure*}[t!]
    \centering
    \includegraphics[width=1.0\linewidth]{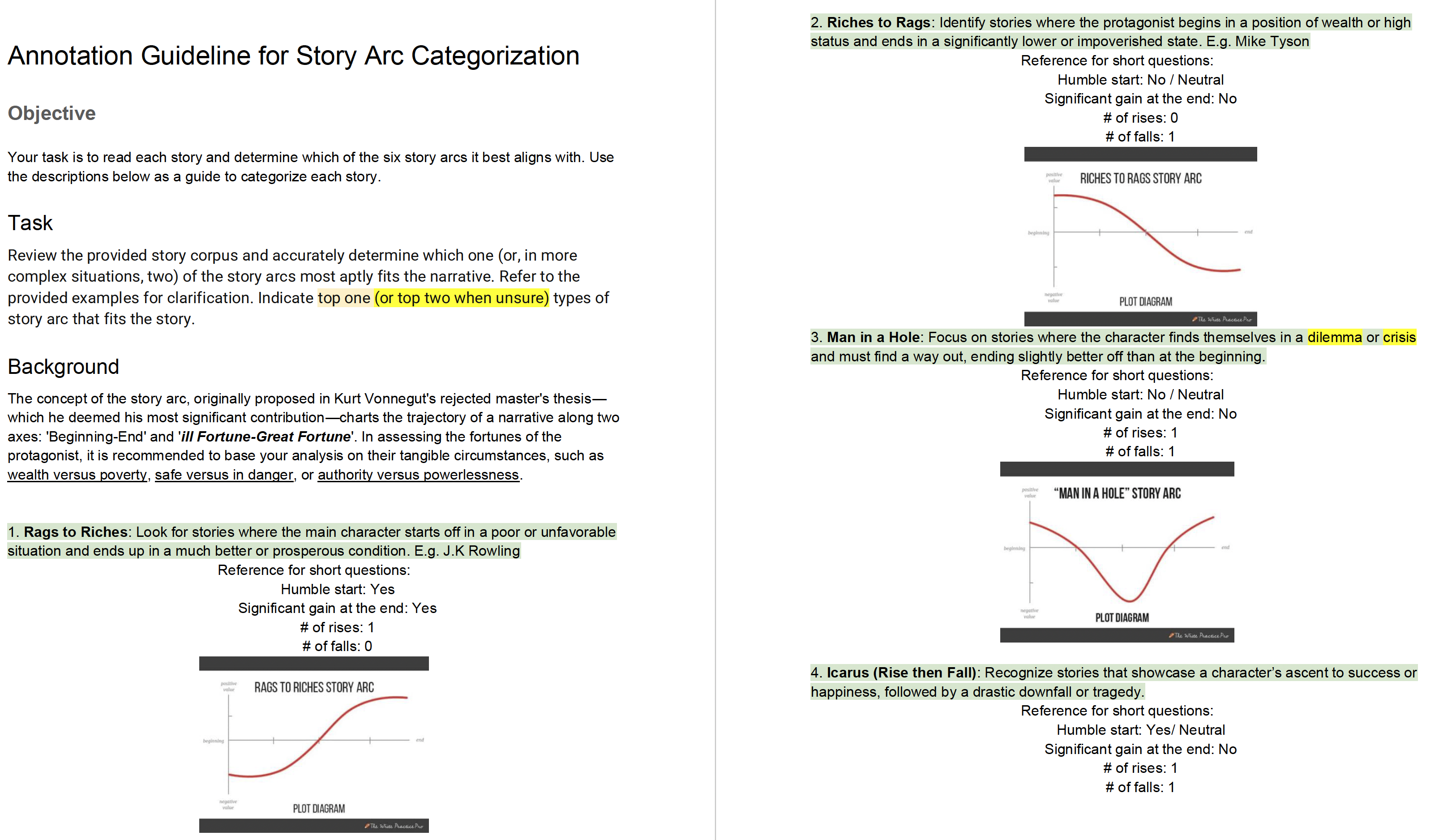}
    \vspace{-2mm}
    \caption{Detailed Annotation Guideline for Story Arc Categorization, Page 1-2.
}
    \label{fig:survey02}
    \vspace{-5mm}
\end{figure*}

\begin{figure*}[t!]
    \centering
    \includegraphics[width=1.0\linewidth]{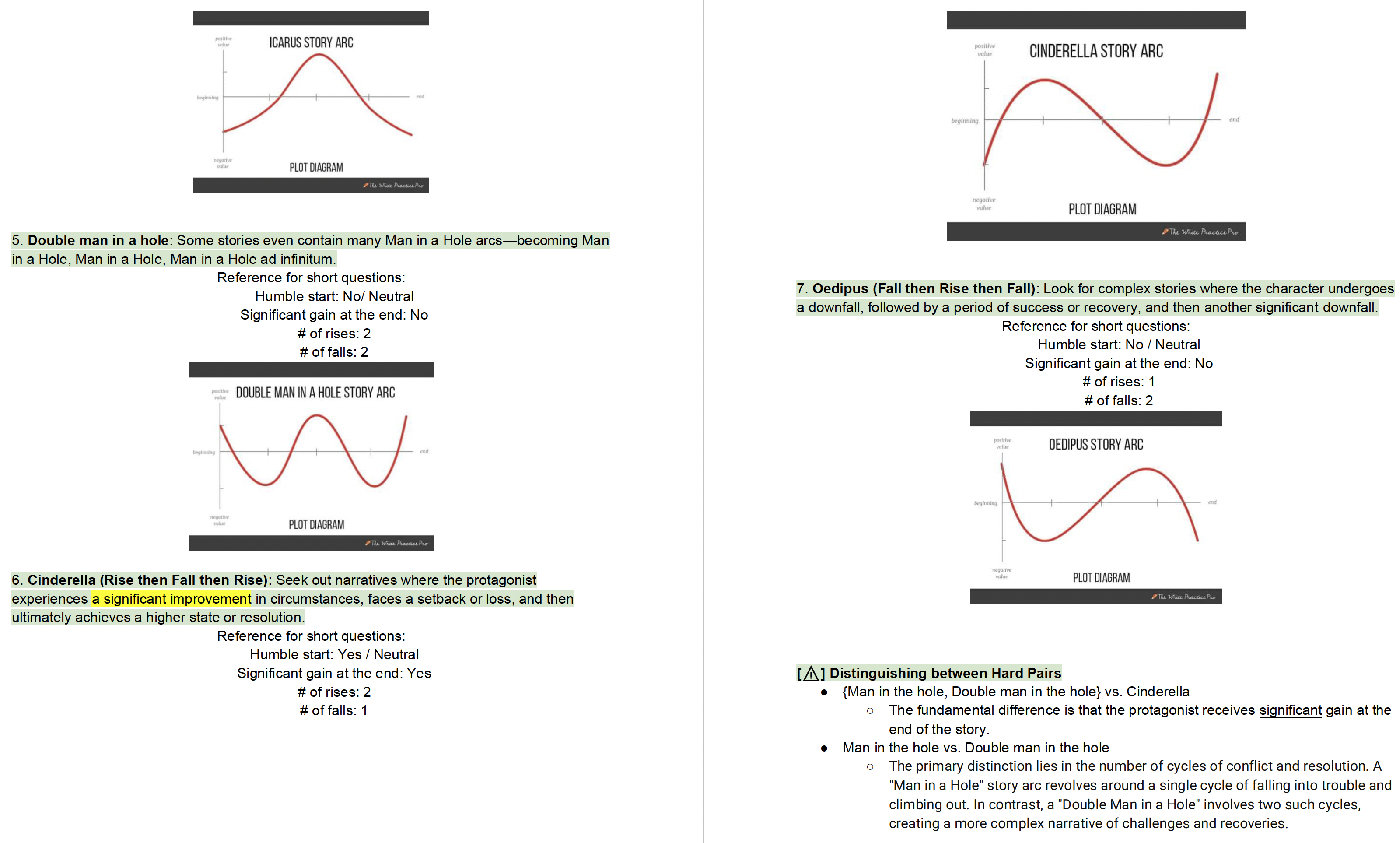}
    \vspace{-2mm}
    \caption{Detailed Annotation Guideline for Story Arc Categorization, Page 3-4
}
    \label{fig:survey03}
    \vspace{-5mm}
\end{figure*}

\begin{figure*}[t!]
    \centering
    \includegraphics[width=1.0\linewidth]{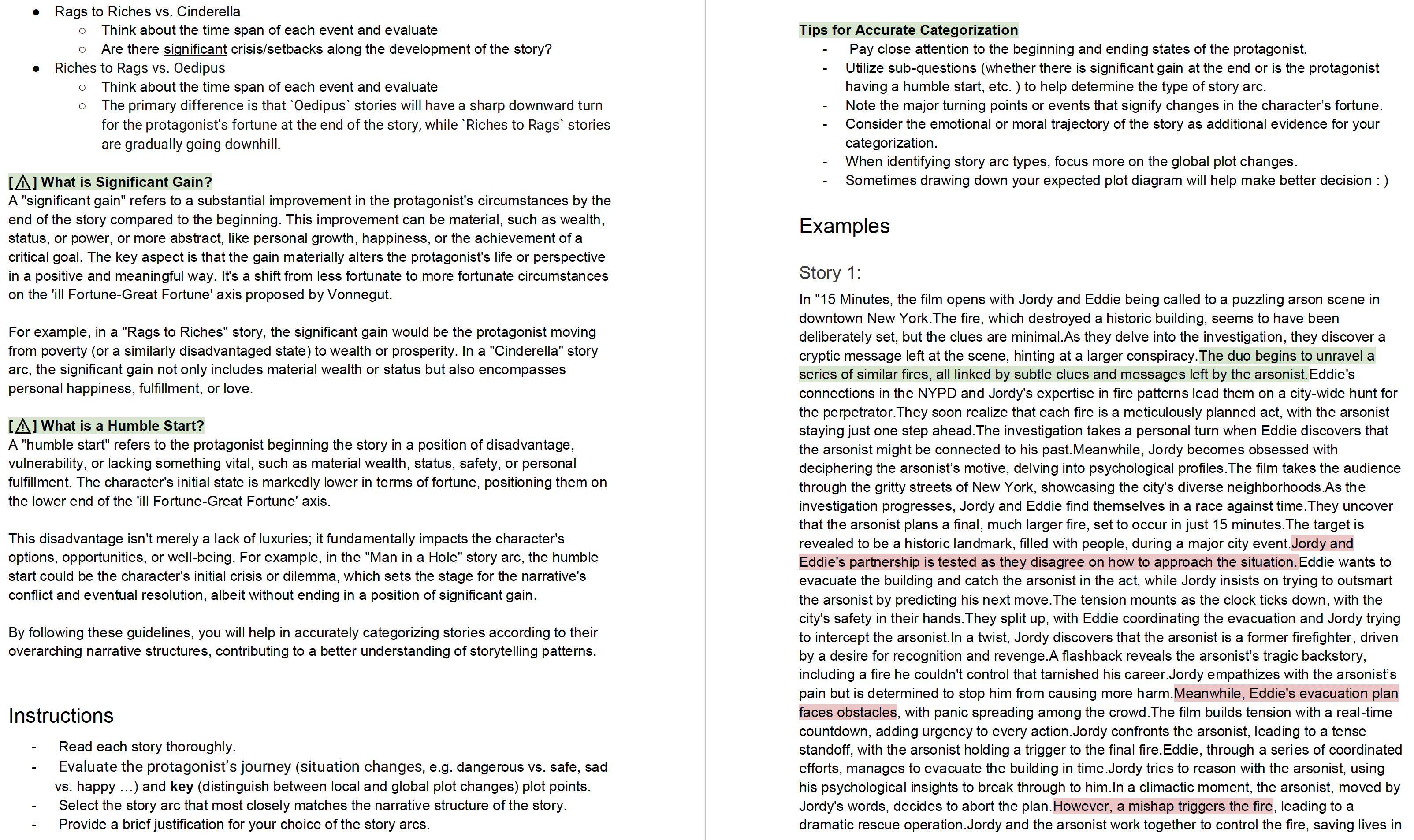}
    \vspace{-2mm}
    \caption{Detailed Annotation Guideline for Story Arc Categorization, Page 5-6.
}
    \label{fig:survey04}
    \vspace{-5mm}
\end{figure*}

\begin{figure*}[t!]
    \centering
    \includegraphics[width=1.0\linewidth]{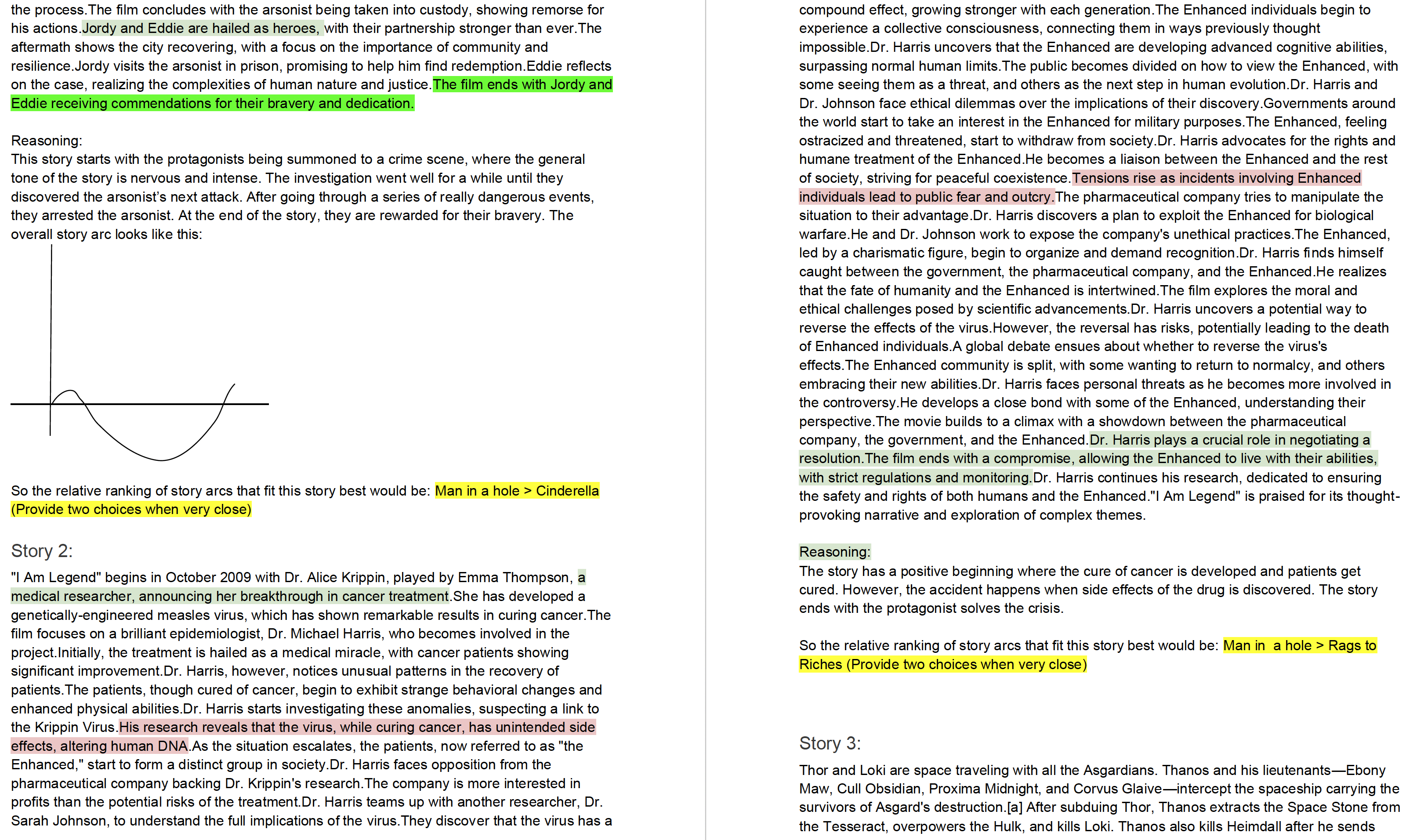}
    \vspace{-2mm}
    \caption{Detailed Annotation Guideline for Story Arc Categorization, Page 7-8.
}
    \label{fig:survey05}
    \vspace{-5mm}
\end{figure*}

\begin{figure*}[t!]
    \centering
    \includegraphics[width=0.95\linewidth]{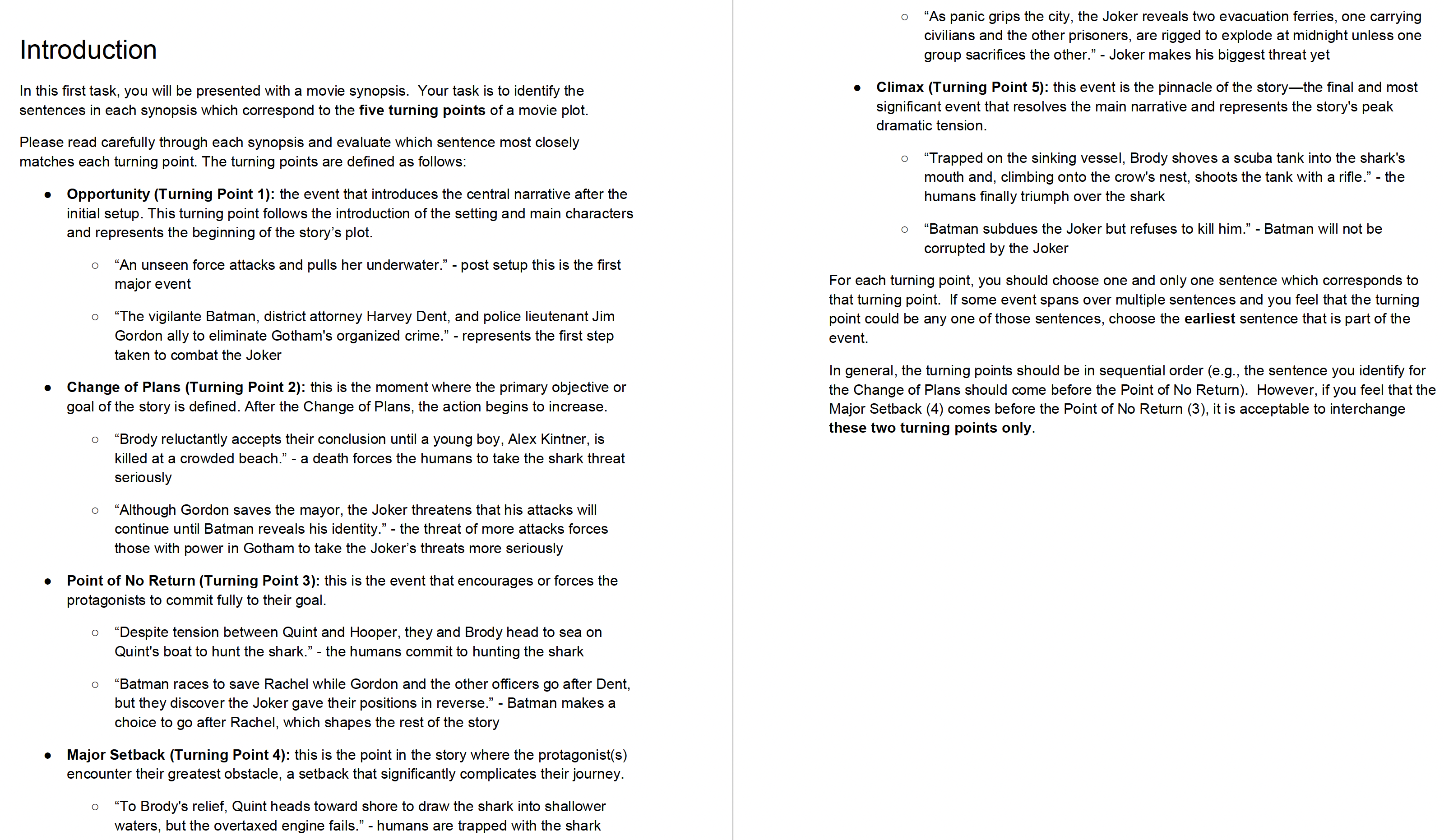}
    \vspace{-2mm}
    \caption{Detailed Annotation Guideline for Turning Point Identification, Page 1-2.
}
    \label{fig:survey06}
    \vspace{-5mm}
\end{figure*}

\begin{figure*}[t!]
    \centering
    \includegraphics[width=0.95\linewidth]{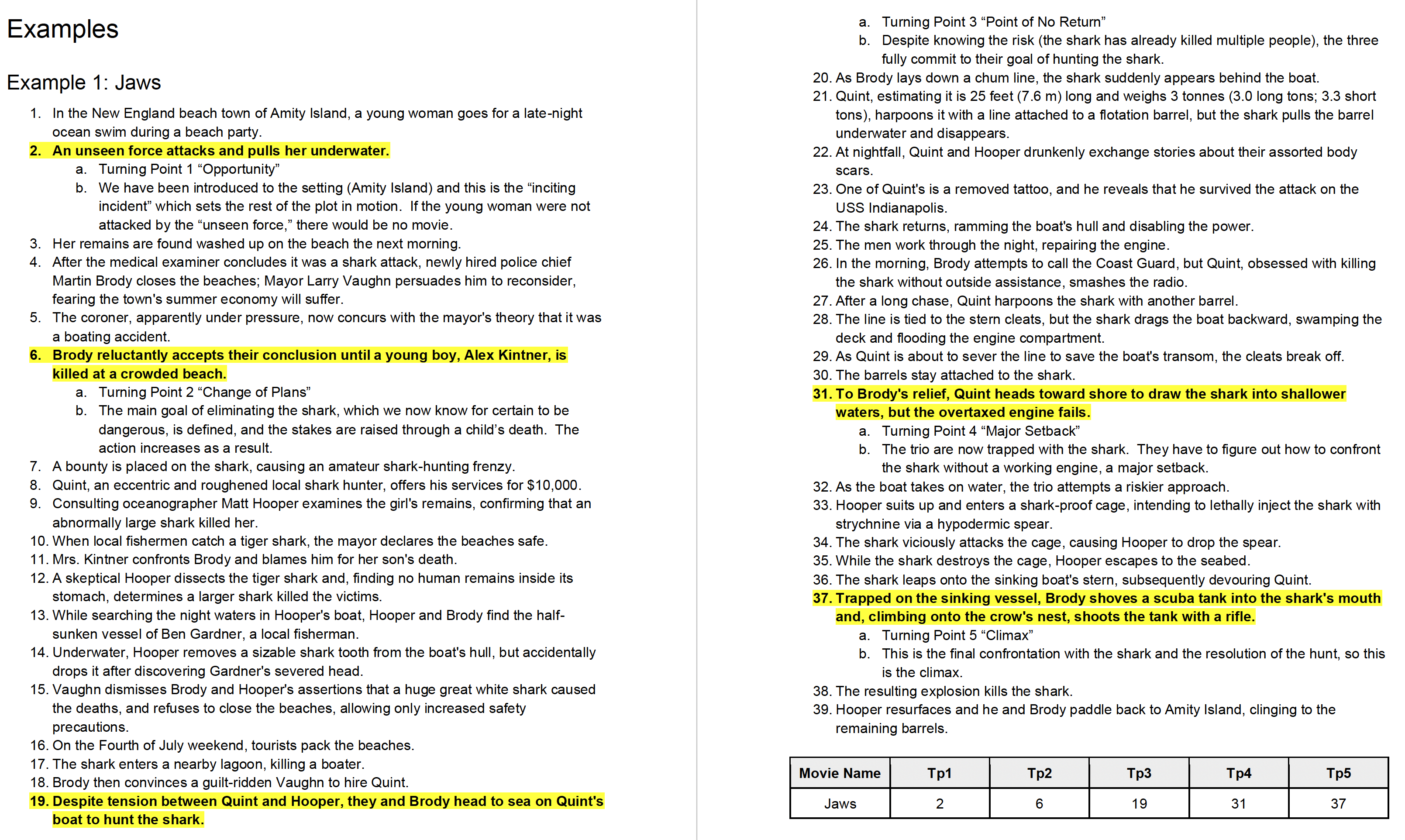}
    \vspace{-2mm}
    \caption{Detailed Annotation Guideline for Turning Point Identification, Page 3-4.
}
    \label{fig:survey07}
    \vspace{-5mm}
\end{figure*}

\begin{figure*}[t!]
    \centering
    \includegraphics[width=0.95\linewidth]{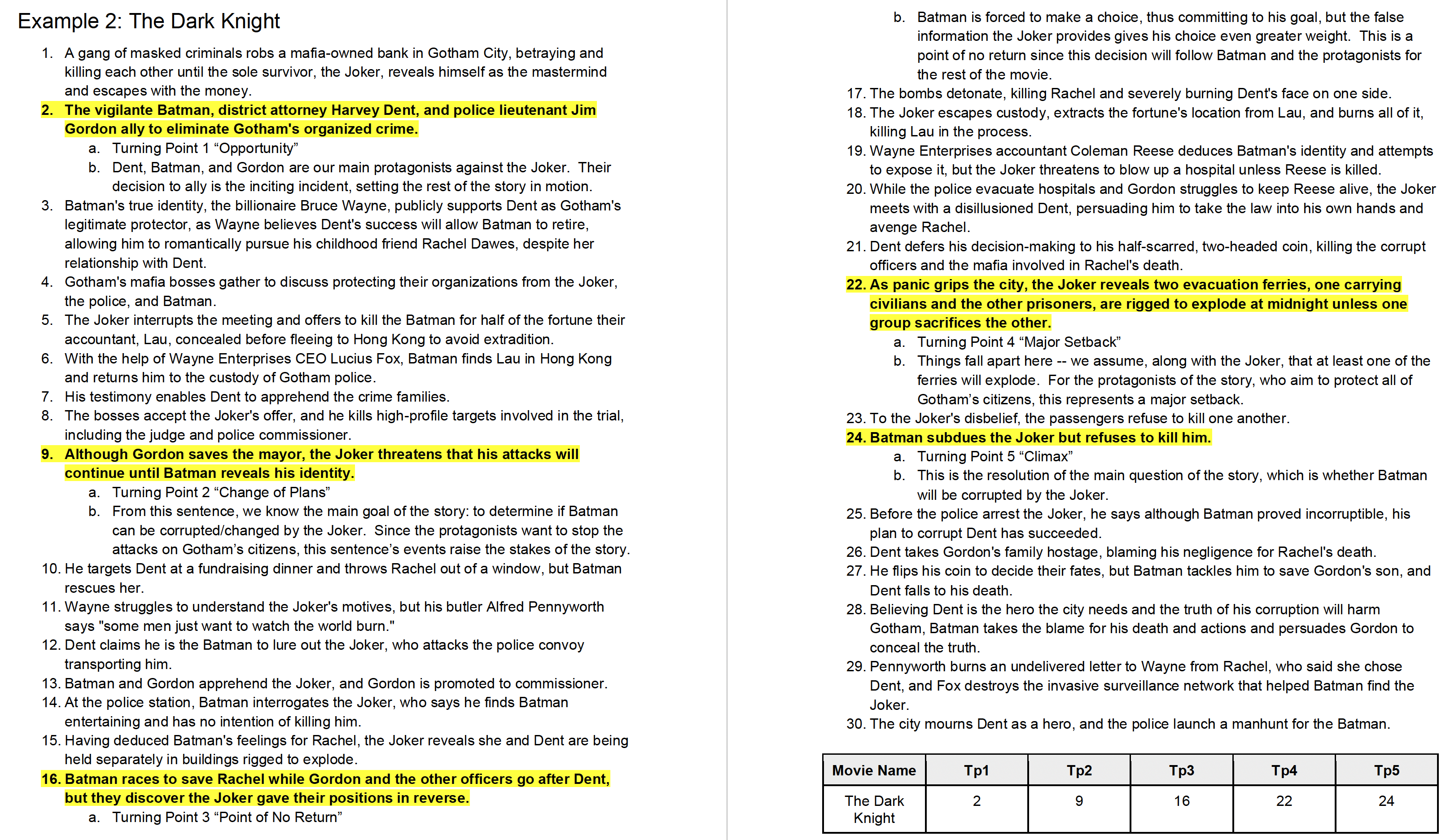}
    \vspace{-2mm}
    \caption{Detailed Annotation Guideline for Turning Point Identification, Page 5-6.
}
    \label{fig:survey08}
    \vspace{-5mm}
\end{figure*}

\begin{figure*}[t!]
    \centering
    \includegraphics[width=0.95\linewidth]{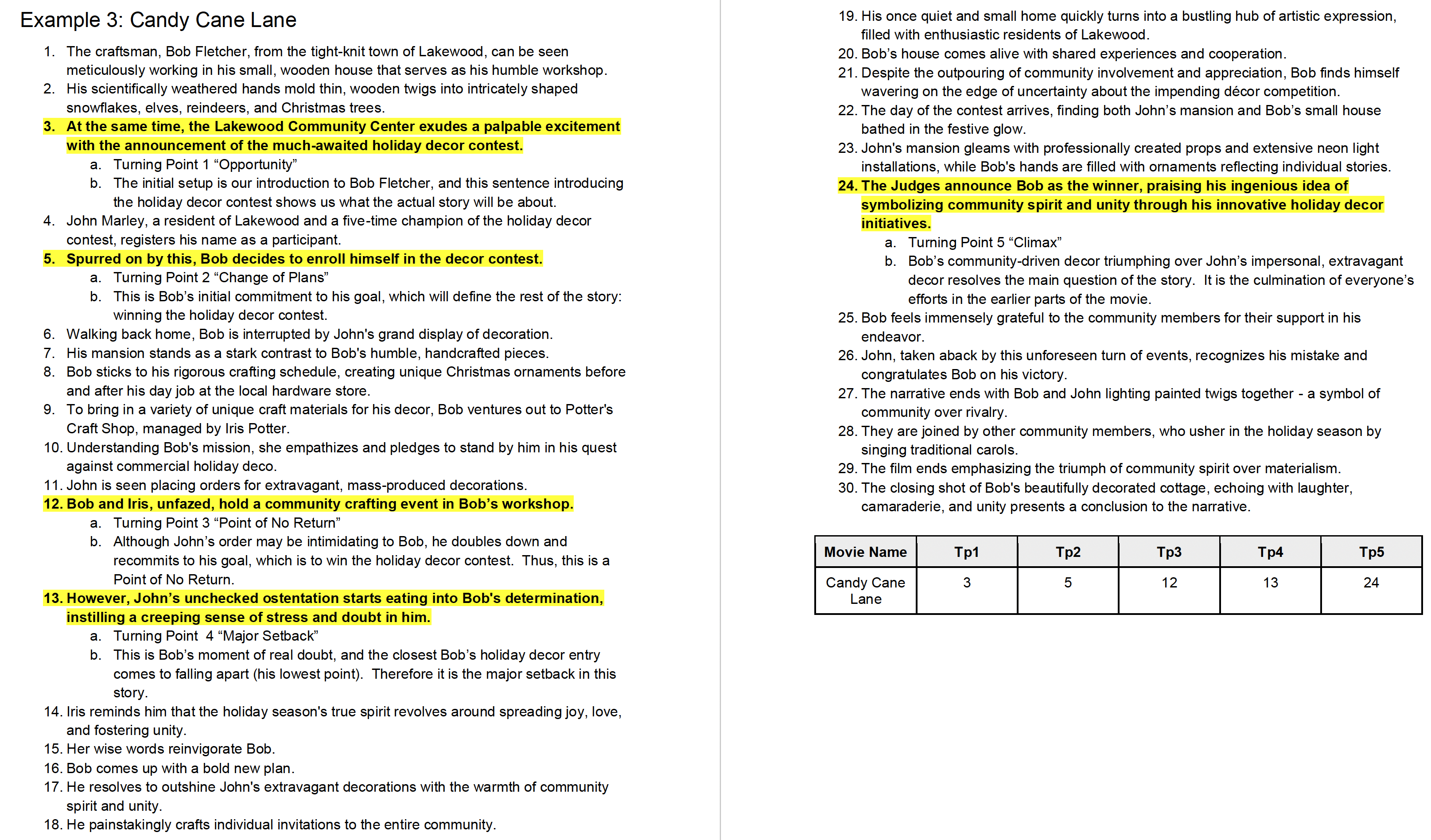}
    \vspace{-2mm}
    \caption{Detailed Annotation Guideline for Turning Point Identification, Page 7-8.
}
    \label{fig:survey09}
    \vspace{-5mm}
\end{figure*}

Figure \ref{fig:survey02} to Figure \ref{fig:survey05} list the detailed annotation guideline and examples of story arc categorization. Figure \ref{fig:survey06} to Figure \ref{fig:survey09} list the detailed annotation guideline and examples of turning point identification.
 
% BEGIN of human annotation interfaces


% END

%\subsection{Benchmark Evaluation} The screenshots of our human evaluation interface for the benchmark experiment can be found in Figure \ref{fig:survey_benchmark_01} and \ref{fig:survey_benchmark_02}.

\begin{figure*}[t!]
    \centering
    \includegraphics[width=0.9\linewidth]{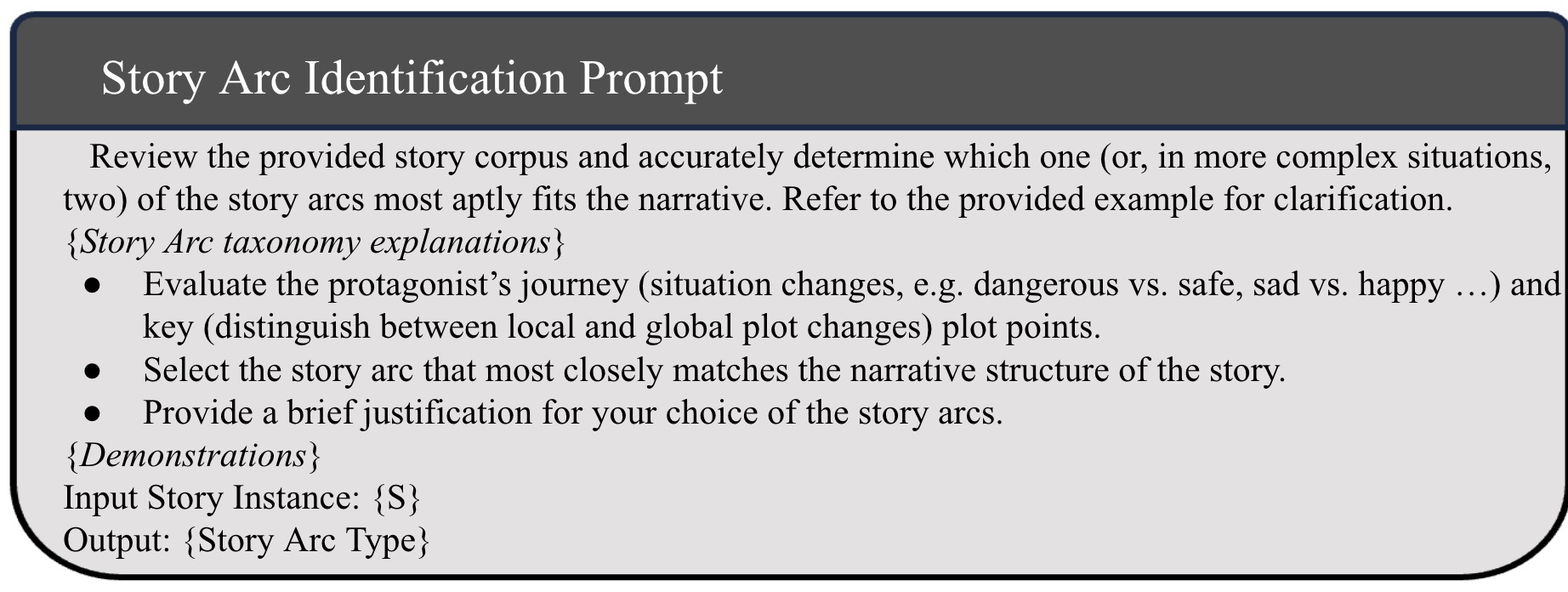}
    \vspace{-3mm}
    \caption{Prompt for Story Arc Identification Task.}
    \vspace{-3mm}
    \label{fig:arc_prompt}
\end{figure*}

\begin{figure*}[t!]
    \centering
    \includegraphics[width=0.9\linewidth]{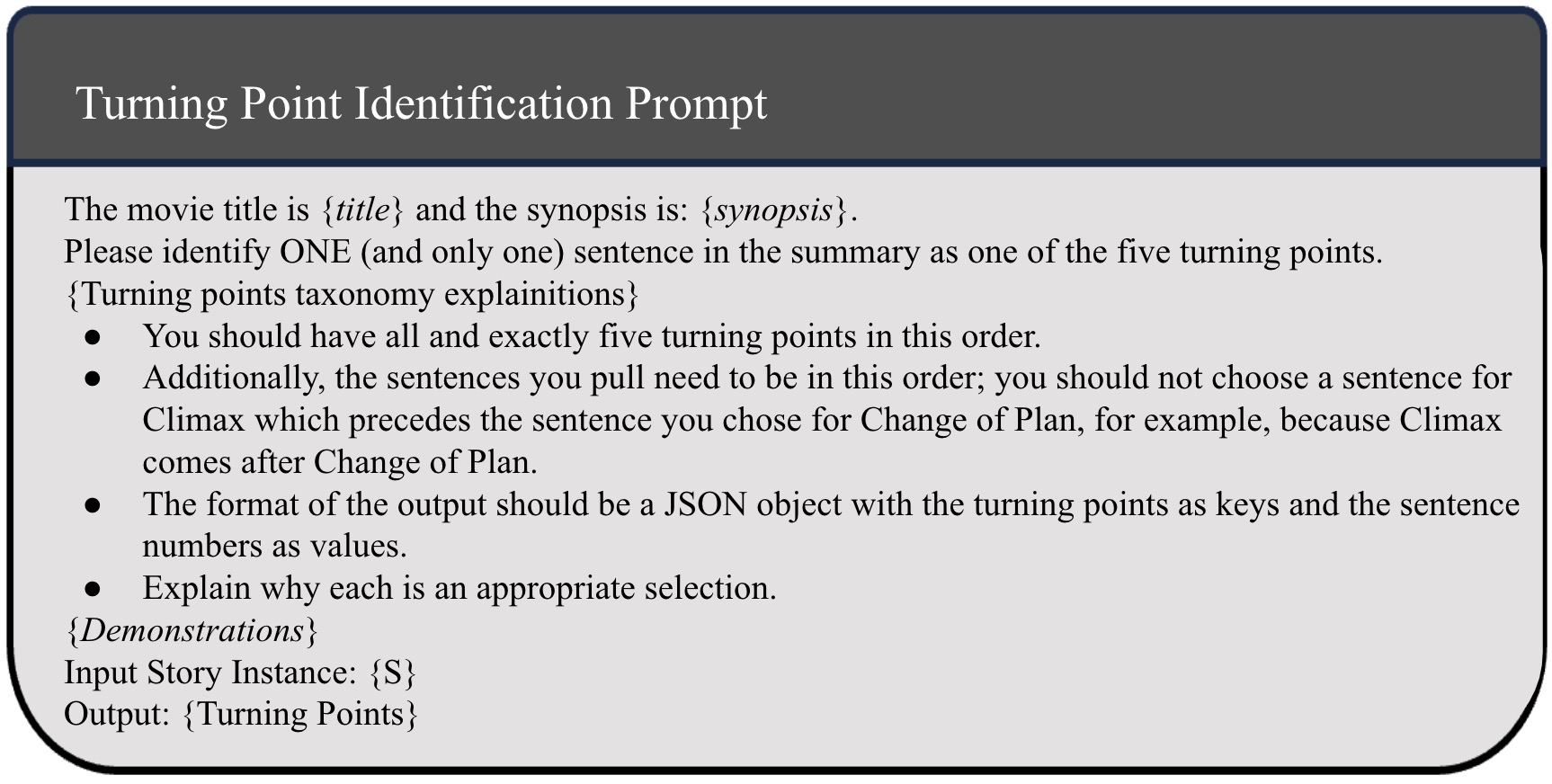}
    \vspace{-3mm}
    \caption{Prompt for Turning Point Identification Task.}
    \vspace{-3mm}
    \label{fig:tp_prompt}
\end{figure*}

\section{Prompt details}

\subsection{Prompts Used in Data Preparation}
We consider the introductory part (first 1-3 sentences in the human-written narrative) as the initial setting. We asked GPT-4 to rephrase the setting by replacing all proper nouns (names, places, anything unique), and then change the phrasing slightly to return the new setting. Based on the newly rephrased initial setting and the original title, we asked the model to generate a similar but not identical title.

\subsection{Prompts Used in Benchmarking}
\label{apdx:prompt_details}
We show the prompt of story arc identification task in Figure \ref{fig:arc_prompt}, and prompt of turning point identification task in Figure \ref{fig:tp_prompt}.

\label{sec:appendix}

\end{document}